\documentclass[sigconf]{acmart}
\usepackage{hyperref}
\usepackage{booktabs} 
\usepackage{subfigure}
\usepackage{xcolor}

\usepackage{changepage}
\usepackage{listings}

\usepackage{float}

\usepackage[ruled]{algorithm2e} 

\SetAlFnt{\small}
\SetAlCapFnt{\small}
\SetAlCapNameFnt{\small}
\SetAlCapHSkip{0pt}

\usepackage{amsmath,amsfonts,bm}
\usepackage{pifont}

\def\eqref#1{equation~\ref{#1}}

\def\1{\bm{1}}

\def\eps{{\epsilon}}

\def\rva{{\mathbf{a}}}

\def\rvm{{\mathbf{m}}}

\def\rvo{{\mathbf{o}}}

\def\rvs{{\mathbf{s}}}

\DeclareMathAlphabet{\mathsfit}{\encodingdefault}{\sfdefault}{m}{sl}
\SetMathAlphabet{\mathsfit}{bold}{\encodingdefault}{\sfdefault}{bx}{n}

\newcommand{\mc}{\mathcal}

\newcommand{\mcI}{\mc{I}}

\newcommand{\norm}[1]{||#1||}

\newcommand{\tnorm}[1]{\norm{#1}_2}

\newcommand{\set}[1]{\{#1\}}

\newcommand{\jjpar}[1]{\left( #1 \right)}

\copyrightyear{2024}
\acmYear{2024}
\setcopyright{acmlicensed}\acmConference[SIGGRAPH Conference Papers '24]{Special Interest Group on Computer Graphics and Interactive Techniques Conference Conference Papers '24}{July 27-August 1, 2024}{Denver, CO, USA}
\acmBooktitle{Special Interest Group on Computer Graphics and Interactive Techniques Conference Conference Papers '24 (SIGGRAPH Conference Papers '24), July 27-August 1, 2024, Denver, CO, USA}
\acmDOI{10.1145/3641519.3657492}
\acmISBN{979-8-4007-0525-0/24/07}

\begin{document}
\title[SuperPADL: Scaling Language-Directed Physics-Based Control]{SuperPADL: Scaling Language-Directed Physics-Based Control with Progressive Supervised Distillation}

\author{Jordan Juravsky}
\orcid{0000-0003-2080-7074}
\affiliation{%
 \institution{NVIDIA}
 \country{Canada}
 }
\affiliation{
 \institution{Stanford University}
 \country{United States}
}
\email{jordanjuravsky@gmail.com}
\author{Yunrong Guo}
\orcid{0000-0001-7468-6162}
\affiliation{%
 \institution{NVIDIA}
 \country{Canada}
 }
\email{kellyg@nvidia.com}
\author{Sanja Fidler}
\orcid{0000-0003-1040-3260}
\affiliation{%
 \institution{NVIDIA}
 \country{Canada}
}
\affiliation{
 \institution{University of Toronto}
 \country{Canada}
}
\email{sfidler@nvidia.com}
\author{Xue Bin Peng}
\orcid{0000-0002-3677-5655}
\affiliation{%
 \institution{NVIDIA}
 \country{Canada}
}
\affiliation{
 \institution{Simon Fraser University}
 \country{Canada}
}
\email{japeng@nvidia.com}

\begin{teaserfigure}
    \centering
    \includegraphics[width=\textwidth]{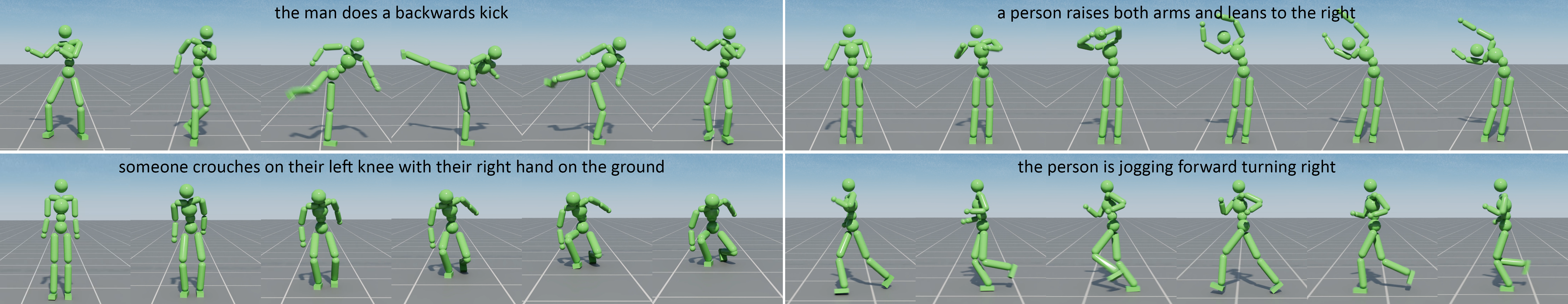}
    \caption{A physically simulated character performing motions specified by language commands. Our framework is able to train a versatile language-directed controller on a large dataset containing thousands of motions.}
    \label{fig:teaser}
\end{teaserfigure}

\begin{abstract}

Physically-simulated models for human motion can generate high-quality responsive character animations, often in real-time. Natural language serves as a flexible interface for controlling these models, allowing expert and non-expert users to quickly create and edit their animations. Many recent physics-based animation methods, including those that use text interfaces, train control policies using reinforcement learning (RL). However, scaling these methods beyond several hundred motions has remained challenging. Meanwhile, kinematic animation models are able to successfully learn from thousands of diverse motions by leveraging supervised learning methods. Inspired by these successes, in this work we introduce SuperPADL, a scalable framework for physics-based text-to-motion that leverages both RL and supervised learning to train controllers on thousands of diverse motion clips. SuperPADL is trained in stages using progressive distillation, starting with a large number of specialized experts using RL. These experts are then iteratively distilled into larger, more robust policies using a combination of reinforcement learning and supervised learning. Our final SuperPADL controller is trained on a dataset containing over 5000 skills and runs in real time on a consumer GPU. Moreover, our policy can naturally transition between skills, allowing for users to interactively craft multi-stage animations. We experimentally demonstrate that SuperPADL significantly outperforms RL-based baselines at this large data scale.

\end{abstract}



\maketitle

\section{Introduction}

Physics-based character animation offers the potential of synthesizing life-like and responsive behaviors from first principles. The use of reinforcement learning techniques for character control has led to a rapid growth in the corpus of motor skills that can be replicated by simulated characters, ranging from common behaviors, such as locomotion \citep{2017-TOG-deepLoco,2020-ALLSTEPS,QuestSim2022}, to highly athletic skills, such as gymnastics and martial arts \citep{2018-TOG-deepMimic,BasketballLiu2018,2PlayerWon2021,2022-Soccer-Juggle}. As the capabilities of simulated characters continue to improve, interfaces that enable users to direct and elicit the desired behaviors from a character will be vital for the viability of these systems for practical applications. The most common interfaces for directing simulated characters often leverage compact control abstractions, such as joystick commands or target waypoints \citep{Treuille2007,Coros09,delasa2010,PFNN2017,2017-TOG-deepLoco}. These abstractions provide accessible interfaces that allow users to easily specify high-level commands for a character, but they tend to offer only limited control over the character's behaviors. More versatile control interfaces, such as target trajectories and keyframes, can in principle allow users to specify any desired behaviors for a character \citep{2018-TOG-deepMimic,wang2020unicon,ScalableWon2020,Luo2023PerpetualHC}. However, authoring target trajectories can itself be a labour-intensive process, requiring significant domain expertise or specialized equipment (e.g. motion capture).

Natural language offers a promising interface that can be both accessible and versatile. Large language models have been shown to provide powerful interfaces for directing generative models in a large variety of domains \citep{bert2018,GPT2020,saharia2022photorealistic,poole2023dreamfusion}. Efforts have also been made to incorporate language interfaces into motion synthesis models. However, the majority of this work has been focused on kinematic motion models \citep{petrovich22temos,dabral2022mofusion,tevet2023human,jiang2023motiongpt,zhang2023generating}. Language-directed controllers for physics-based characters have yet to replicate comparable versatility, scalability, and motion quality of their kinematic counterparts \citep{2022-SA-PADL,ren2023insactor,sun2023prompt}. In this work, we aim to develop a scalable framework for training language-directed controllers for physically simulated characters, which is able to leverage large motion datasets to learn a single versatile controller capable of performing a vast repertoire of skills.

The central contribution of this work is a large-scale framework for training versatile language-directed controllers for physically simulated characters. To address scalability challenges of applying reinforcement learning to a large corpus of skills, we propose a progressive distillation framework, which starts from skill-specific controllers, and then progressively constructs more versatile controllers that are able to perform a larger and larger set of skills, ultimately yielding a single unified controller capable of reproducing behaviors from over 5000 motion capture sequences. Our model operates in real-time, and is able to respond interactively to changes in the user's language commands.

\section{Related Work}

Physics-based models for character animation has had a long history in computer graphics, with a large body of work devoted to constructing controllers that enable simulated characters to reproduce a large repertoire of motor skills \citep{AthleticsHodgins1995,Silva2008SimulationOH,BipedWang2009,DataDrivenLee2010,BicycleTan2014,MuscleWang2012,BasketballLiu2018,DressClegg2018}. While these efforts have led to substantial improvements in the motor capabilities of simulated characters, a barrier that has precluded the wider adoption of these models for practical applications has been the lack of accessible interfaces for users to direct the behaviors of these models. The majority of these models provide users with simple control interfaces, such as joystick commands or target waypoints \citep{Treuille2007,Coros09,MotionFields2010,taskBasedLocomotion2016,PFNN2017,2018-TOG-deepMimic,NSM2019,AmorphousZhang2020,2020-TOG-MVAE,2021-TOG-AMP,TimeCritical2021,Lee2021Parameterized,2022-TOG-ASE}, which are easy to use but greatly restrict the control a user can exert over the character's behaviors. Alternatively, motion imitation models provide an expressive interface, where a user can specify target reference motions for a simulated character to imitate \citep{2018-TOG-deepMimic,DreCon2019,ScalableWon2020,Luo2023PerpetualHC}. While these models can provide users with versatile and granular control over a character's behavior, they require users to construct reference motions for every desired motion, which can itself be a costly and labour intensive process.

\paragraph{Text-Driven Generation:} Natural language offers the potential to develop accessible and expressive interfaces for directing the behaviors of simulated characters. Recent advances in large-language models has lead to a proliferation of text-driven interfaces for generative models in a wide variety of domains \citep{bert2018, robertaLiu, t5Raffel,saharia2022photorealistic,poole2023dreamfusion}. Similar techniques have also been adopted for motion synthesis, creating text2motion models that are able to generate motions according to natural language descriptions \citep{PlappertMA17,linvigil18,Language2Pose2019,MotionClipTevet2022,tevet2023human}. These models have in part been made possible by the availability of large public datasets which contain tens of hours of text-labeled human motion data \citep{HumanML3D2022, BABEL:CVPR:2021}. A large body of work has also adapted generative modeling techniques for text-directed motion synthesis, such as sequence prediction models with RNNs \cite{PlappertMA17,linvigil18}, variational auto-encoders \citep{petrovich22temos,TEACH:3DV:2022}, contrastive coding \citep{Language2Pose2019,MotionClipTevet2022}, transformers \citep{jiang2023motiongpt,zhang2023motiongpt}, and diffusion models \citep{dabral2022mofusion,tevet2023human}. While these text2motion models have demonstrated promising capabilities, the vast majority of this work focuses on kinematic motion models.

\paragraph{Language-Directed Controllers:} In this work, we aim to develop a large-scale framework for training language-directed models for physically simulated characters. Prior efforts in this domain have largely been limited in terms of the variety of motions that can be reproduced by a model, and the quality of the generated motions. \citet{2022-SA-PADL} proposed an adversarial imitation learning framework for training language-directed controllers for a simulated humanoid character. However, their system was only effective in learning from a relatively small dataset of approximately 9 minutes of motion data. A common approach for developing more versatile controllers is to combine a kinematic text2motion model with a low-level motion tracking model \citep{ren2023insactor,Luo2023PerpetualHC}. These methods first generate a kinematic reference motion using previously mentioned text2motion techniques, and the task of controlling a simulated character is then reduced to simple motion tracking. This approach is able to leverage the scalability of kinematic motion models that are trained on large motion datasets, but the capabilities of the simulated character are then greatly restricted to closely follow the behaviors dictated by the kinematic model. Furthermore, the motion quality of these composite models is still conspicuously lower compared to state-of-the-art physics-based character animation systems.

The goal of our work is to develop versatile end-to-end language-directed controllers for physically simulated characters that can be controlled using natural language to perform a large corpus of motor skills. Our model does not require an existing kinematic text2motion model, and instead directly maps language commands to motor actuations for driving a simulated character to perform the behaviors specified by the user. The key component of our system is a progressive supervised distillation framework, which gradually trains controllers on larger and larger datasets. Training specialized expert controllers has been a common approach for applying motion imitation methods to train general controllers from large datasets \citep{ScalableWon2020,Luo2023PerpetualHC}. However, these prior systems combine individual experts into a general controller by constructing mixture-of-experts models, which requires retaining all experts in the runtime model. Our framework leverages a progressive distillation procedure to progressively aggregate expert controllers trained on different motion datasets into a single general controller capable of reproducing behaviors from a large dataset of motion clips. Distillation has been applied in previous efforts to scale up reinforcement learning to train multi-task control policies \citep{merel2018neural,CatchCarry2020}. \citet{wagener2023mocapact} used a single-stage distillation approach to train a general motion tracking controller capable of imitating approximately 3.5 hours of motion data. Our multi-stage progress distillation approach enables our system to train versatile controllers on 8.5 hours of text-labeled motion clips, leading to a unified end-to-end controller that can be directed to perform a large variety skills with simple text commands. Model-based RL has also shown promise for scaling to larger motion datasets \citep{PhysVAEWon2022,ControlVAEYao2022}. However, these techniques have yet to be effectively demonstrated on large motion datasets. Our work shows that model-free RL combined with a progressive distillation approach can be an effective method for training a general unified controller from thousands of motion clips.

\section{Overview}

In this work, our goal is to train a single versatile control policy capable of responding to thousands of different text commands, while being able to naturally transition between motions. The approach we propose, which we call SuperPADL\footnote{The name is inspired by our combination of \textbf{super}vised learning and PADL-like adversarial RL objectives.}, is inspired by two observations:
\begin{itemize}
    \item Existing reinforcement learning techniques for physics-based animation are able to train policies with many desirable properties, such as the ability to reproduce reference motions with high quality and naturally transition between skills. However, these approaches do not scale beyond at most several hundred motions.
    \item Kinematic motion models, such as motion diffusion models, are able to scale to datasets containing thousands of motions using supervised learning objectives.
\end{itemize}

In light of these observations, we present a method that combines RL and supervised objectives, centered around the progressive distillation of motion controllers. Initially, we seek to train a large number of highly specialized expert policies using RL. We then iteratively distill these experts together with supervised techniques, progressively training more general-purpose and capable models. Concretely, our method is composed of three training stages:

\begin{enumerate}
    \item We first train an independent expert tracking policy on every motion capture sequence in our dataset using DeepMimic \citep{2018-TOG-deepMimic}. The purpose of this stage is to create high-quality reconstructions of our original data in the physical domain. These reconstructions provide us with a ``dataset of actions'' that we can use to train later networks with supervised losses.

    \item Next, we randomly partition our dataset into groups of 20 motions and train a controller policy on each group. These group controllers are trained using a hybrid objective that combines adversarial RL with a behaviour cloning (BC) loss on trajectories from the DeepMimic experts. Each group controller learns to naturally transition between the skills in its group, and does not require the phase variable used to train the DeepMimic experts. Note that these group controllers are not conditioned on language during training.

    \item Finally, we distill the group controllers into a single text-conditioned global policy. This final distilled controller is trained exclusively with a supervised imitation objective: for every motion in our dataset, we encourage the distilled controller to match the actions of the appropriate group controller. 

\end{enumerate}

\begin{figure}
    \centering
    \includegraphics[width=0.7\columnwidth]{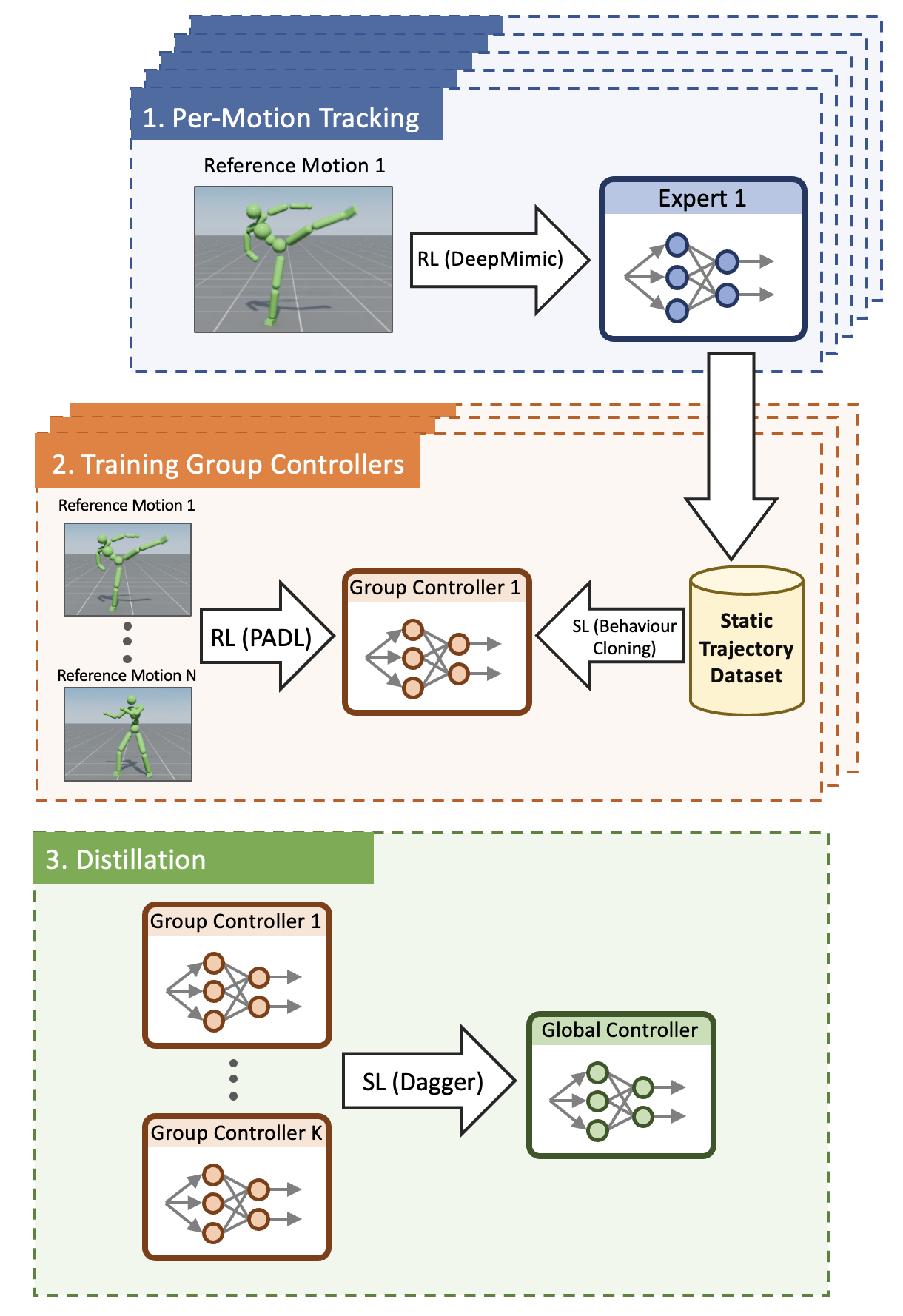}
    \caption{An overview of the SuperPADL training process.}
    \label{fig:overview}
\end{figure}

Figure~\ref{fig:overview} provides an overview of our framework. In each stage of training, we progressively produce more general-purpose, capable controllers. Simultaneously, our approach progressively reduces the use of RL as the number of motions learned per policy grows: DeepMimic experts are trained purely with RL, group controllers are trained with a hybrid RL-BC objective, and the global controller is only trained with supervision. The key to our approach is to only leverage RL methods at the smaller data scales where these methods are effective, and later then transfer the skills of many smaller networks into larger ones through distillation.

\subsection{Training Per-Motion Expert Tracking Policies}
\label{sec:tracking}

Recent work has assembled large datasets $D = \set{(m_i, C_i)}$ of motion capture sequences $m_i$ annotated with a set of one or more natural language labels $C_i$ \citep{BABEL:CVPR:2021, HumanML3D2022}. However, these motion capture recordings are kinematic, preventing the direct application of supervised losses when training physically-simulated control policies. 
In order to address this incompatibility, in the first stage of SuperPADL we translate our motion capture dataset into the physical domain by training an expert tracking policy on every motion in our dataset. 

Our approach is inspired by the first stage of MoCapAct \cite{wagener2023mocapact}, leveraging DeepMimic as our tracking method \citep{2018-TOG-deepMimic}. 
We train a DeepMimic expert policy $\pi^e_i(\rva_t | \rvo_t, \phi)$ on every motion capture sequence in our dataset, conditioned on the current state of character $\rvo_t$ as well as a phase variable $\phi \in [0,1]$ that synchronizes the policy to the reference motion.

We train each policy for a maximum of 3000 epochs, corresponding to approximately 200M frames of experience. To reduce overall compute usage, we monitor the cartesian pose error throughout training and stop training early when the error falls below 3cm. If the epoch limit is reached and the pose error remains above 5cm, we discard the motion from our dataset.
By leveraging a GPU-accelerated implementation of DeepMimic in NVIDIA Isaac Gym \citep{2018-TOG-deepMimic, IsaacGym2021}, the majority of tracking policies complete training in under an hour, while only 5\% of motions are discarded (see Section~\ref{sec:tracking-results}). The discarded items in our dataset often correspond to physically implausible motions, such as those that involve third-party objects that are not recorded in the motion capture sequence (e.g. climbing up nonexistent stairs, sitting on nonexistent chairs, etc.). Additionally, our tracking environment resets whenever a character's body part, excluding the feet, touches the ground. This causes motions like crawling and lying prone to be discarded.

After training each tracking policy, we record a dataset of trajectories containing 10000 observation-action frames from each expert:
\begin{align}
    T_i &= (\rvo_1^i, \rva_1^i, \rvo_2^i, \rva_2^i, ... ).
\end{align}
In practice, in order to increase the diversity of states encountered during these rollouts, $T_i$ is created in chunks by initializing the character 100 different times using random frames from the reference motion and rolling out each initialization for 100 frames.
Additionally, in 90\% of rollouts, stochastic actions are sampled from the policy's action distribution, while in the remaining 10\% deterministic/greedy actions are taken from the mean of the action distribution. Note that while a combination of stochastic and deterministic actions are used to generate the rollouts, the actions recorded in the trajectory data always correspond to the deterministic action that the expert policy would have taken at each state. This is done to avoid adding noise to the trajectory action labels.
We refer to this collection of trajectories $D_T = \set{T_i}$ as our trajectory dataset.

\subsection{Training Group Controllers with PADL+BC}
\label{sec:group}

Our ultimate goal is to train a single model that can perform a wide range of behaviors in response to user text commands and seamlessly transition between different behaviors as the user's command changes. 
However, at the end of the first stage of SuperPADL training, we are instead left with a collection of highly specialized expert tracking policies. Each expert can reproduce their corresponding motion capture clip with high quality, yet they cannot generate any other motion. Additionally, tracking policies lack the robustness to reliably recover from perturbations or initialization in out-of-distribution states, as might be necessary when transitioning between different motions.
Therefore, the next step in our pipeline is to develop more general control policies that retain the motion quality of individual experts while also being much more robust.

Previous work has demonstrated that adversarial reinforcement learning can effectively train policies that exhibit these properties \citep{2021-TOG-AMP, 2022-TOG-ASE, 2022-SA-PADL}. However, as we experimentally demonstrate in Section~\ref{sec:global-results}, these RL techniques do not directly scale to thousands of motions. 
We address this limitation by employing a progressive distillation approach in the second and third stages of SuperPADL training. 
In the second stage of SuperPADL, we train controllers on small groups of motions using an objective combining adversarial RL and behaviour cloning, before performing a second, purely-supervised distillation stage to train a final global policy. 
In this first distillation stage, we randomly partition our dataset into groups of 20 motions:

\begin{align}
    P_i &= \set{(m_{20i + 1}, C_{20i + 1}), (m_{20i + 2}, C_{20i + 2}), ... , (m_{20i + 20}, C_{20i + 20})}
\end{align}

and train a group controller $\pi^g_i(\rva_t | \rvo_t, I)$ on each partition $P_i$, parameterized by the current character state $\rvo_t$ and the motion index $I \in \set{20i + 1, ..., 20i + 20}$. The index $I$ is encoded using a trainable, randomly-initialized embedding table.

Our goal is for each group controller to:

\begin{itemize}
    \item Imitate the motions in its partition when conditioned on the corresponding motion.
    \item Naturally transition between these motions when the input index changes.
    \item Generally avoid falling over.
\end{itemize}
We optimize each group controller using PADL+BC, a novel objective that combines the adversarial RL setup of PADL with behaviour cloning: 
\begin{align}
    \mathcal{L} &= \mathcal{L}_{\text{PADL}} + 0.01\mathcal{L}_{\text{BC}} .
\end{align}
PADL introduces a motion-conditioned discriminator network:
\begin{align}
    \text{Disc}(I, \rvs,\rvs') \to [0, 1]
\end{align}
which is trained to distinguish between state transitions $(\rvs, \rvs')$ from a reference motion $m_I$ and state transitions generated by the policy when conditioned on $I$. 
The policy is optimized with PPO \citep{PPO2017}, using a reward that encourages the policy to ``fool'' the discriminator: 
\begin{align}
    r_t &= -\mathrm{log}\left(1 - \text{Disc}(I, \rvs_{t-1}, \rvs_{t}) \right).
\end{align}
In addition to this PPO loss, we introduce an additional behaviour cloning loss on the  dataset of expert trajectories $D_T$. At each step of optimization, we sample (observation, action) pairs from the stored trajectories of our grouped motions and encourage the group controller to imitate the expert actions:
\begin{align}
    \mathcal{L}_{\text{BC}} &= \mathop{\mathbb{E}}_{I \sim \set{20i + 1, ..., 20i + 20}} \mathop{\mathbb{E}}_{(\rvo, \rva) \sim T_I} || \pi^g_i(\rvo, I) - \rva ||_2^2.
\end{align}
While the observations for the tracking experts included a phase variable to synchronize the policy to the target motion, this phase variable is omitted from the observations given to the group controller. This is crucial since determining the correct phase observation to give to the group controller during inference is difficult - for example, when transitioning from one motion to another, it is unclear what the phase should be set to. 

The motivation behind the added behaviour cloning loss is twofold. First, the supervised training signal allows us to significantly cut down on the training cost of a group controller relative to a pure-PADL policy. We train each group controller using a 2000-epoch warmup phase where only the BC loss is applied. These warmup epochs complete significantly faster than normal PPO epochs where trajectories must be rolled out in the environment. Following this warmup, we only train group controllers with PPO+BC for an additional 1B samples of experience, compared to the 7B samples reported in \citet{2022-SA-PADL} for training a PADL policy from scratch. Overall, training a group controller with PADL+BC completes in around 12 hours on a single A40 GPU, compared to almost three days of training for a pure-PADL controller. Additionally, we demonstrate in Section~\ref{sec:group-results} that PADL+BC controllers, leveraging the experts' accurate motion reconstructions, can generate motions with higher quality than pure PADL policies.

\subsection{Distilling into a Global Text-Conditioned Policy}
\label{sec:global}

Group controllers mark a significant improvement in generalization over the tracking expert policies, enabling a single controller to reproduce multiple motions, naturally transition between motions, and operate without a phase variable. However, the group controllers are still constrained by the relatively small set of motions that each controller is trained on. As we demonstrate in Section~\ref{sec:global-results}, PADL+BC alone does not scale to the thousands of motions available in open-source motion capture datasets \citep{mahmood2019amass}. Moreover, our group controllers are conditioned using simple motion indices and cannot follow commands in natural language.

To train a global, language-conditioned policy $\pi^G(\rva_t | \rvo_t, c)$ (where $c$ denotes a text caption), we perform a second round of distillation, now leveraging the group controllers as teacher policies. Unlike PADL+BC, which combines RL and supervised training objectives, the training of the global policy is purely supervised, allowing us to scale to much larger datasets. 

Training of the global policy begins similarly to group controllers with an offline, behaviour cloning warmup phase using the static trajectory dataset $D_T$ collected from the tracking policies. This initializes the state distribution of the global policy to be similar enough to those of the group controllers that they can provide effective teacher feedback. 

Following warmup, the global policy is trained to convergence using online imitation learning similar to DAGGER \citep{ross2011reduction}. Every epoch, trajectories are rolled using the current global controller. Each observation in those trajectories is then annotated using the appropriate group controller (based on the motion that the policy is trying to imitate). The global controller is optimized to match these annotations.
Unlike with group controllers, when conditioning the global controller to imitate motion $i$, we provide a natural language caption for the motion instead of simply an index variable. At every rollout, we sample a motion $(m_i, C_i)$ from $D$ and sample a caption $c$ from $C_i$. The caption is encoded using the CLIP text encoder, with the pooled encoder embedding provided to the main policy network $\pi^G$.

\subsection{Experimental Details}

\subsubsection{Dataset Curation}

The motion capture data that we use to train SuperPADL is a filtered subset of AMASS, an open-source aggregation of smaller motion datasets \citep{mahmood2019amass}. We first filter out any motion clips shorter than two seconds or longer than nine seconds. We also apply a series of filters that attempt to detect motions that are physically impossible, such as climbing a staircase or swinging from a bar (third party objects are not included in the motion capture recordings). The plausibility filters examine the heights of the character's limbs and extremities to look for signs that a character has not touched the ground for a prolonged period of time, filtering the motion out of our dataset if such an event is detected. We train DeepMimic experts on a dataset of 5866 filtered motions.

To augment this motion data with natural language annotations, we use the HumanML3D dataset of captions \citep{HumanML3D2022}, which provides several captions for every motion in AMASS. To add additional diversity to this data, we use ChatGPT to generate paraphrases using the original set of annotations. When training our global controller on the 5587 motions that pass the expert tracking phase (totalling approximately 8.5 hours of data), there are a total of 48207 captions in the dataset.

\subsubsection{Network Architectures}

\begin{figure*}
    \centering
    \subfigure
    [Expert Controller]{\includegraphics[width=0.2\textwidth]{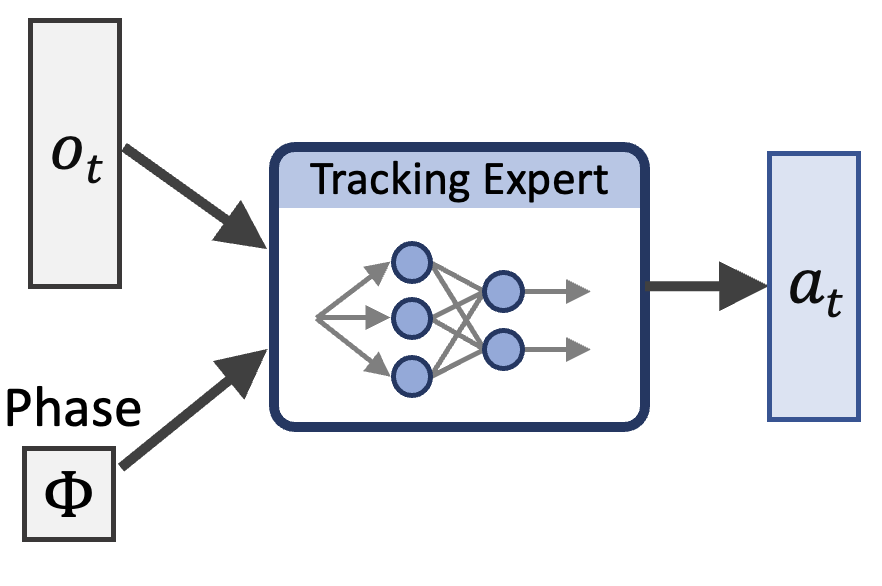}}
    \hspace{0.2cm}
    \subfigure[Group Controller]
    {\includegraphics[width=0.2\textwidth]{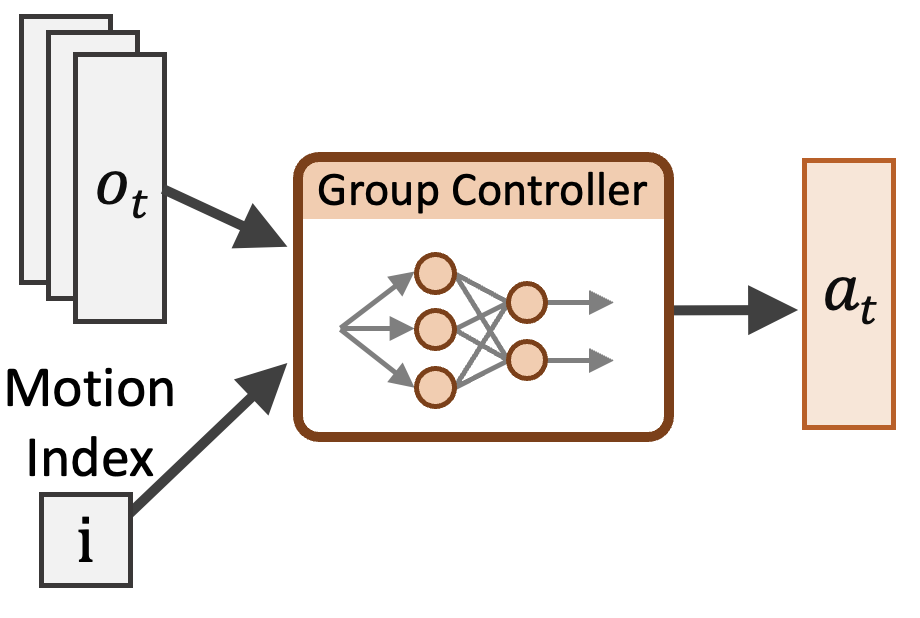}}
    \hspace{0.2cm}
    \subfigure[Global Controller]
    {\includegraphics[width=0.36\textwidth]{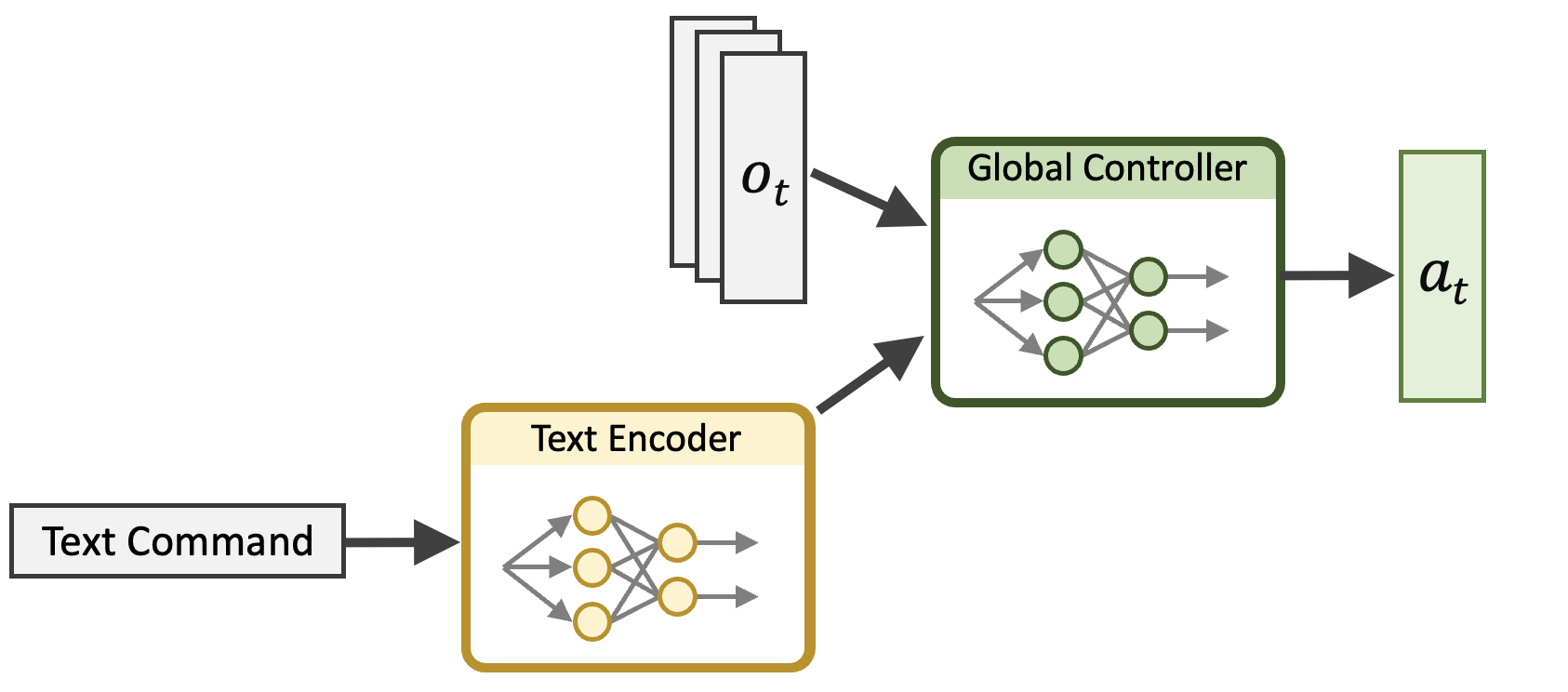}}\\
    \vspace{-0.4cm}
    \caption{Network architectures for controllers at each stage of SuperPADL. All controllers are modelled with simple MLPs.}
    \label{fig:networks}
\end{figure*}

All policies (tracking experts, group controllers, and global controllers) are trained using simple MLP architectures. Figure~\ref{fig:networks} summarizes the inputs and outputs of each network. While several existing works in adversarial RL use only the character's state at the the current frame as model observations \citep{2021-TOG-AMP, 2022-TOG-ASE, 2022-SA-PADL}, we observe benefits when training group and global controllers that are conditioned on a longer history. We maintain a context window looking 40 frames back into the past, and generate inputs for actor and critic networks by selecting every eighth frame from the window, totalling five total frames of observations. This approach balances providing models with a longer history with restricting the total observation dimension, since excessively large observations can slow down and potentially destabilize training. 
More architectural details are given in Appendix~C.

\section{Results}

\begin{figure*}[t]
	\centering
    \subfigure[``a person is walking backwards slowly'']{\includegraphics[width=0.49\textwidth]{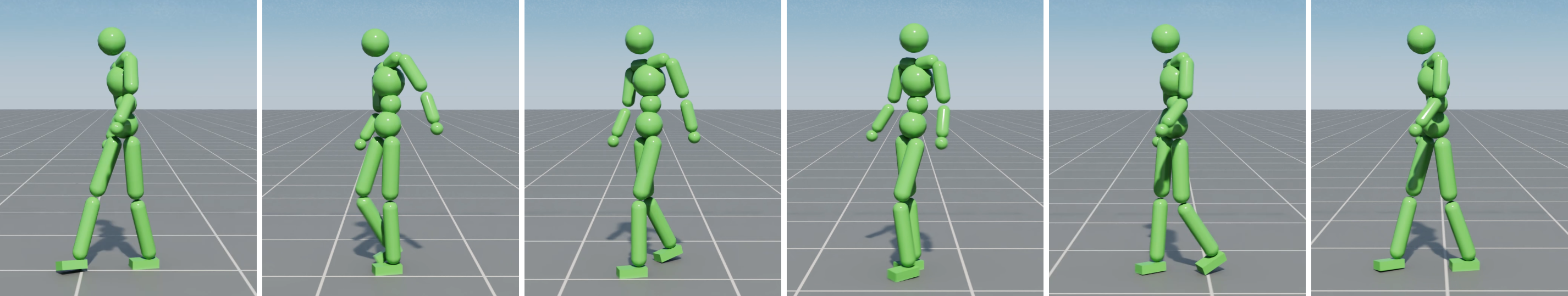}}
    \subfigure[``the person is jogging lightly'']{\includegraphics[width=0.49\textwidth]{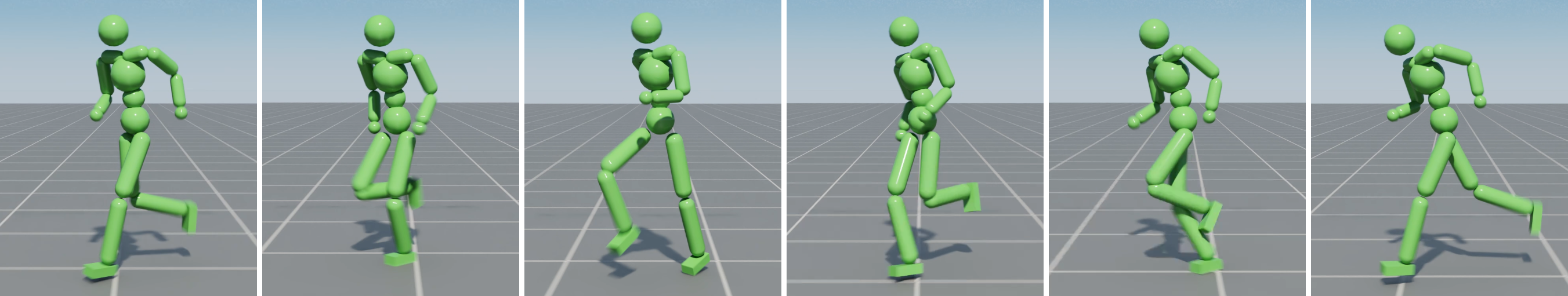}} \\
    \vspace{-0.2cm}
    \subfigure[``a man crouches down on the ground'']{\includegraphics[width=0.49\textwidth]{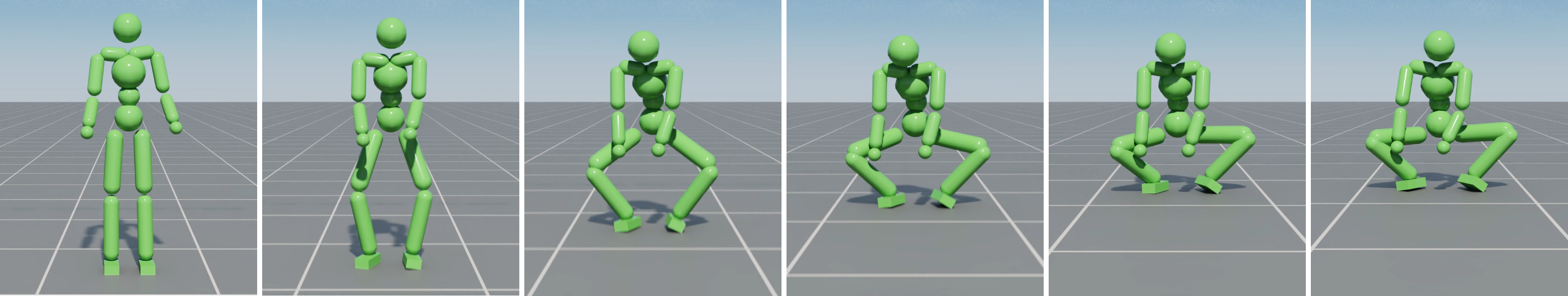}} \subfigure[``a person is doing star jumps'']{\includegraphics[width=0.49\textwidth]{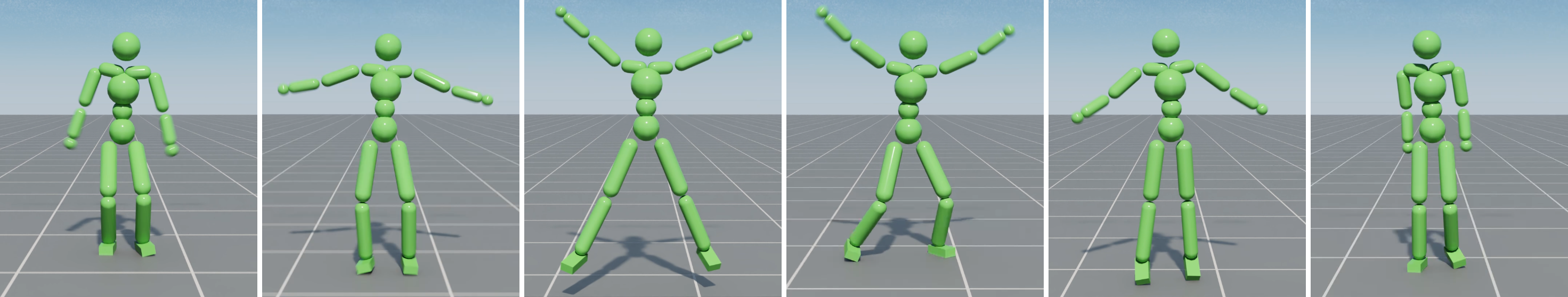}} \\
    \vspace{-0.2cm}
    \subfigure[``a man does a kick to the side'']{\includegraphics[width=0.49\textwidth]{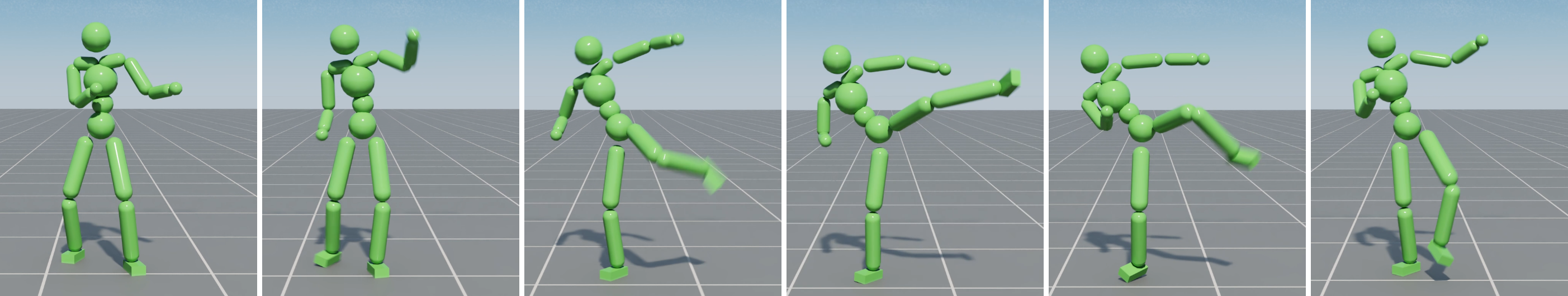}} 
    \subfigure[``in a fighting stance, person punches downward with their left hand'']{\includegraphics[width=0.49\textwidth]{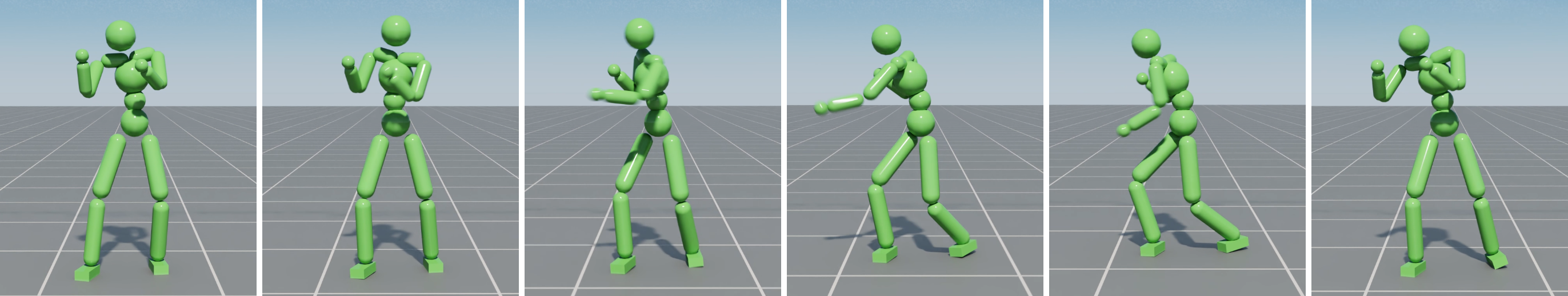}} \\
    \vspace{-0.2cm}
    \subfigure[``a person rotates with his hands on his hips like they're doing a hullahoop'']{\includegraphics[width=0.49\textwidth]{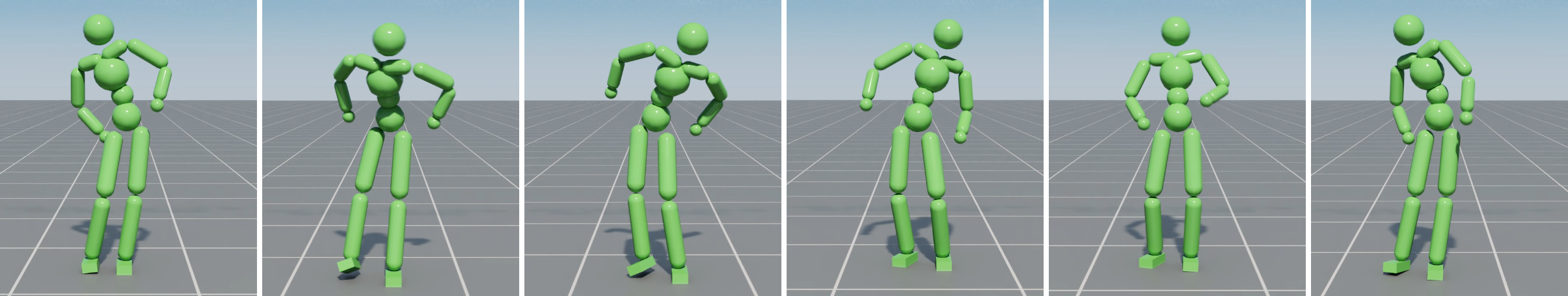}}
    \subfigure[``a person dances and moves around with their hands in the air'']{\includegraphics[width=0.49\textwidth]{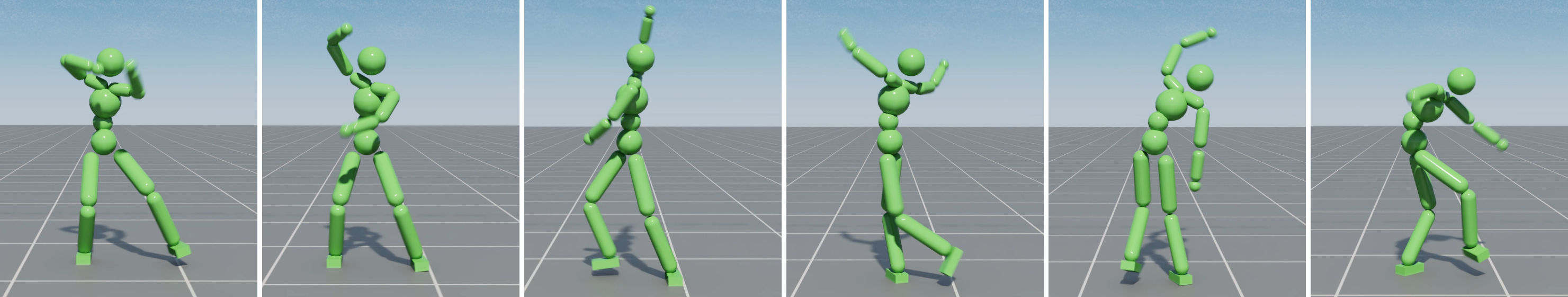}} \\
    \vspace{-0.2cm}
    \subfigure[``a man throws then catches an object'']{\includegraphics[width=1\textwidth]{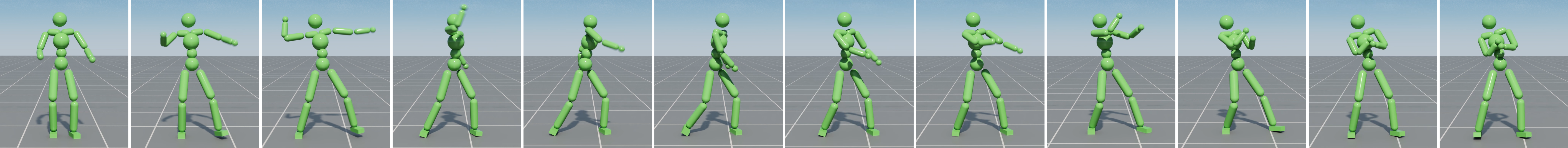}}\\
    \vspace{-0.2cm}
    \caption{Simulated character performing skills specified by language commands. Our framework is able to train a single text-conditioned controller that can perform a diverse array of skills.}
    \label{fig:filmstrips}
\end{figure*}

We present examples of our global controller reproducing motions from its training data using text commands in Figure~\ref{fig:filmstrips}. We demonstrate that SuperPADL is able to generate an extremely diverse set of motions, ranging from basic locomotion skills and hand gestures to much more difficult martial arts and dancing behaviours. In contrast with kinematic motion diffusion models, where generating a single animation can take up to a minute \citep{tevet2023human}, we highlight that SuperPADL can generate motion in real time on a single consumer GPU, enabling interactive applications.

\begin{figure*}[t]
	\centering
    \subfigure[``a person is walking backwards slowly'' $\rightarrow$ ``the man is sprinting happily'']{\includegraphics[width=1\textwidth]{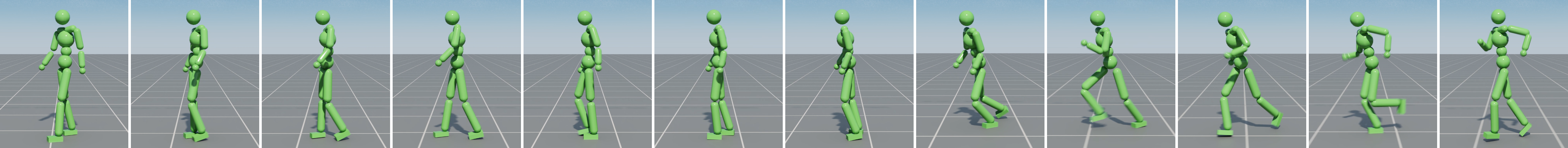}}\\
    \vspace{-0.2cm}
    \subfigure[``someone crouches on their left knee with their right hand on the ground and their left arm lifted behind them'' $\rightarrow$ ``the man does a backwards kick'']{\includegraphics[width=1\textwidth]{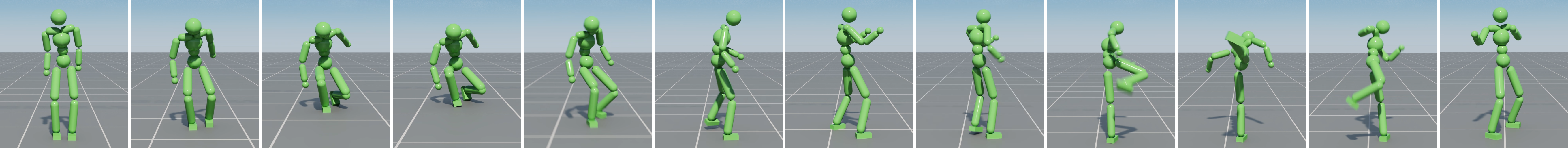}}\\
    \vspace{-0.2cm}
    \subfigure[``the person puts their hand on their face then squats down like they're going underwater'' $\rightarrow$ ``the person is flailing their arms around''.]{\includegraphics[width=1\textwidth]{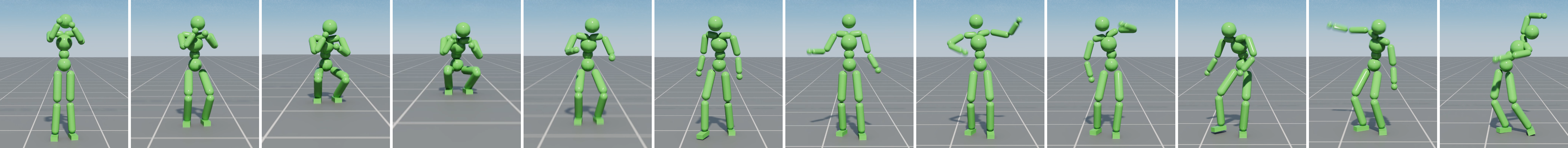}}\\
    \vspace{-0.2cm}
    \caption{Our controller can respond to changing commands in real-time and transition between different skills.}
    \label{fig:filmstrips-transitions}
\end{figure*}

SuperPADL is also able to successfully transition between skills, with examples shown in Figure~\ref{fig:filmstrips-transitions}. These transition abilities were initially learned by each group controller using adversarial RL, and have been inherited by the global controller through distillation. Note that even though each group controller was only trained to transition between the 20 motions in its group, the global SuperPADL controller is able to transition between any two motions, regardless of their group assignments.

\subsection{Training Tracking Experts}
\label{sec:tracking-results}

\begin{figure}
    \centering
    \includegraphics[width=0.8\columnwidth]{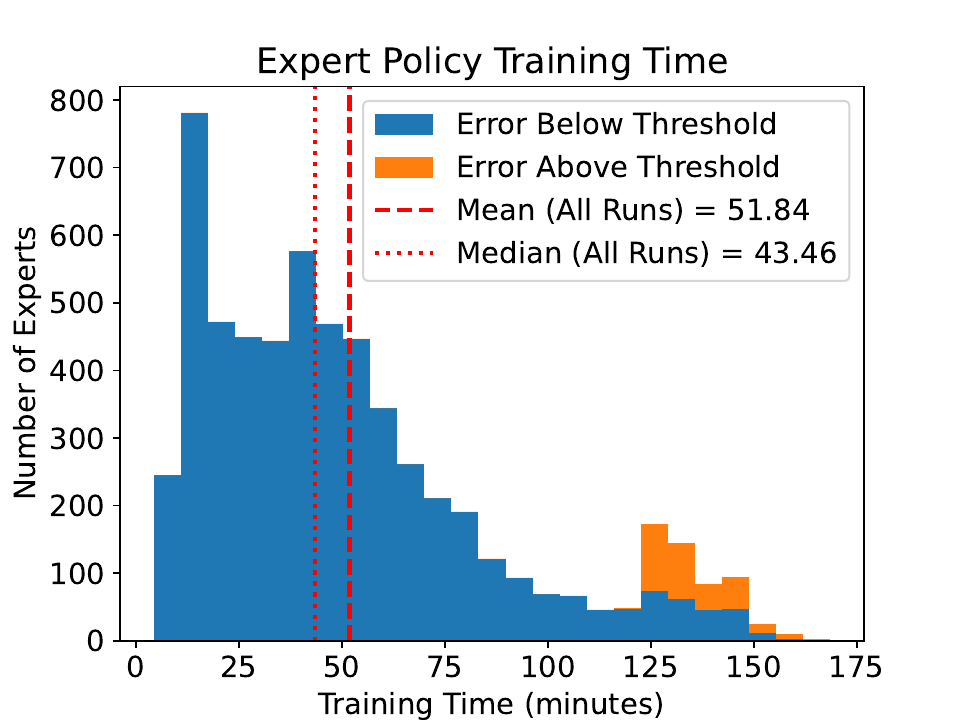}
    \caption{The distribution of training times for tracking experts. The majority of policies terminate training early in less than an hour upon attaining a sufficiently low tracking error.}
    \label{fig:tracking-times}
    \vspace{-0.6cm}
\end{figure}

In Figure~\ref{fig:tracking-times} we visualize the distribution of training times for the expert tracking policies detailed in Section~\ref{sec:tracking}. All policies were trained on individual NVIDIA A40 GPUs. We see that ending training early based on the most recent tracking error significantly reduces the total cost of compute required to train all experts. A majority of experts finish training in under an hour, and over 30\% complete in less than 30 minutes. However, since policies that do not reach the target error threshold are trained until the epoch limit of 3000 epochs, we see that the 5\% of rejected policies have an oversized impact on cumulative training cost. This highlights the importance of strong dataset filters that can identify physically-implausible motions before training begins.

\subsection{Measuring Controller Quality with Thresholded Precision and Recall}
\label{sec:define-metrics}

Measuring tracking error is difficult for policies lacking a phase variable input that synchronizes them to a reference motion. In order to evaluate the quality of motions produced by non-tracking policies (i.e. group and global controllers), we introduce the metrics of thresholded precision and recall. These metrics are inspired by the thresholded coverage metric used in \citep{2022-SA-PADL}, with our thresholded recall metric being almost identical to that work's construction of thresholded coverage. Thresholded recall measures the fraction of a reference motion that is reproduced by a policy when conditioned to generate that motion. To calculate the recall of a policy $\pi$ on a motion sequence $\hat{\rvm} = (\hat{\rvs_0}, \hat{\rvs_1}, ..., \hat{\rvs_n})$, we first roll out a (deterministic) trajectory $\tau = (\rvs_0, \rvs_1, ..., \rvs_k)$ from $\pi$. We then consider all ten-frame-long sliding windows from $\hat{\rvm}$ and check whether any ten-frame window in $\tau$ is ``sufficiently close'', as determined by some threshold $\eps$. Specifically, we define:
\begin{align}
    \text{Rec}(\tau, \hat{\rvm}, \eps) = \frac{1}{n-9} \sum_{i = 0}^{n - 10} \mcI \jjpar{ \jjpar{ \min_{j \in \set{0, ..., k - 10}} \tnorm{\hat{\rvs}_{i:i+9} - \rvs_{j:j+9}} } \le \eps}
    \label{eqn:recall}
\end{align}
where $\rvs_{x:y}$ denotes the concatenation of frames $(\rvs_x, \rvs_{x+1}, ..., \rvs_{y})$, and $\mcI$ denotes an indicator variable. The key difference between this metric and thresholded coverage is that thresholded recall operates on windows of consecutive states, while thresholded coverage only considers individual frames. We choose to construct windows to better capture the temporal structure of the reference motion: for example, a hypothetical trajectory that perfectly imitated $\hat{\rvm}$ in reverse would always produce a perfect thresholded coverage score, but this is not true for thresholded recall.

Complementing this thresholded recall metric is a thresholded precision metric, which considers all the windows in $\tau$ and measures whether any window in $\hat{\rvm}$ is sufficiently close:
\begin{align}
    \text{Prec}(\tau, \hat{\rvm}, \eps) = \frac{1}{k-9} \sum_{i = 0}^{k - 10} \mcI \jjpar{ \jjpar{ \min_{j \in \set{0, ..., n - 10}} \tnorm{\rvs_{i:i+9} - \hat{\rvs}_{j:j+9}} } \le \eps}
    \label{eqn:precision}
\end{align}
While thresholded recall measures the fraction of $\hat{\rvm}$ that the policy imitates, thresholded precision measures the fraction of $\tau$ that imitates a portion of $\hat{\rvm}$. For example, a trajectory that perfectly loops a subset of the reference motion would score very highly on precision, but low on recall. Conversely, a trajectory that perfectly imitated the entire reference clip, but then contained some bizarre additional motions, would have a very high recall score but a very low precision. In practice, we find that the two metrics are correlated, however policies that mostly ignore their conditioning and focus on staying upright (such as the global controller baselines in Section~\ref{sec:global-results}) will often have a higher precision score than recall.

To evaluate trained policies, we follow the procedure of \citet{2022-SA-PADL} and sweep over many values of $\eps$ when calculating thresholded precision and recall. We record trajectories and calculate metrics for all motions in the training dataset, and report the averaged results as a plot. Additionally, to summarize these plots with individual scalars, we report the area under each curve (AUC).

\subsection{Evaluating Global Controllers}
\label{sec:global-results}

\begin{figure}
    \centering
    \subfigure
    {\includegraphics[width=0.95\columnwidth]{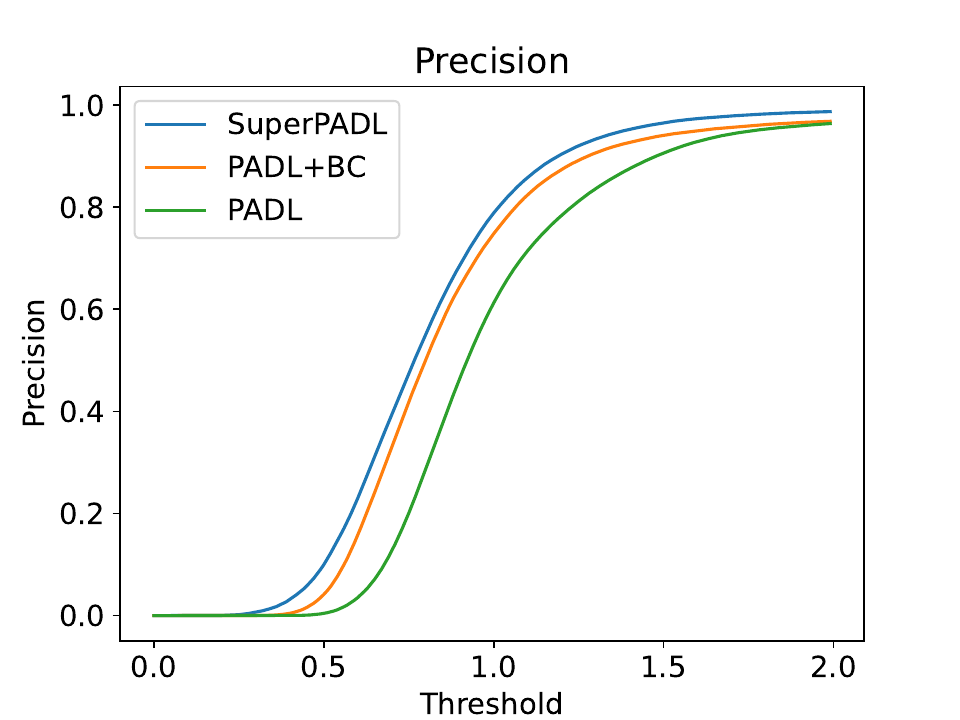}}
    \subfigure
    {\includegraphics[width=0.95\columnwidth]{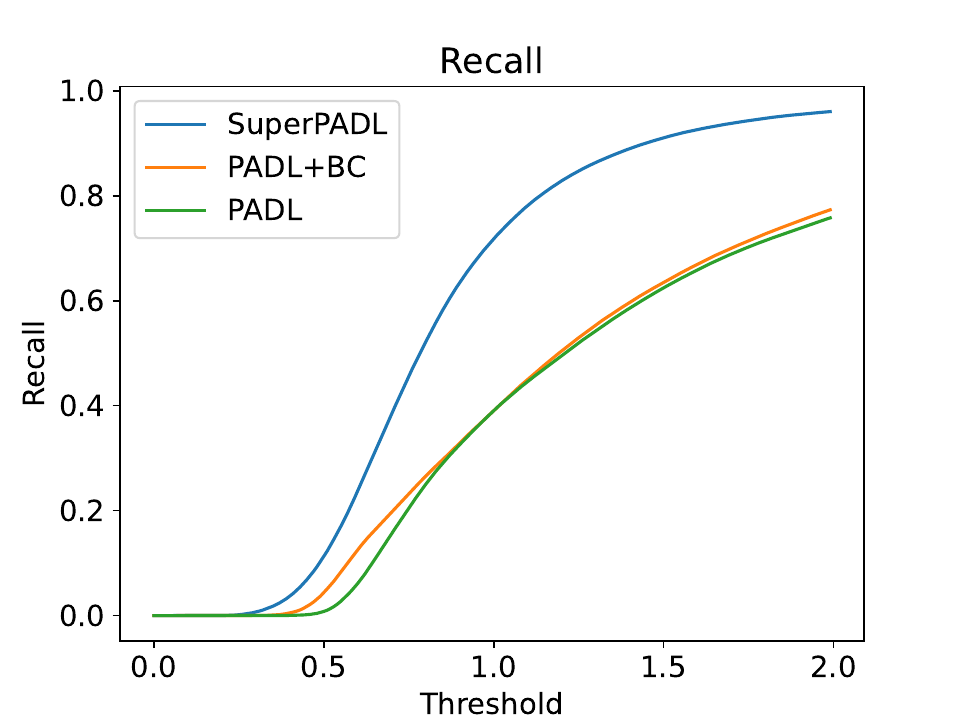}}
    \caption{Thresholded precision and recall metrics for the SuperPADL global controller as well as PADL and PADL+BC baselines. We observe that the SuperPADL global controller has consistently higher precision and recall.}
    \label{fig:global-metrics}
\end{figure}

We use thresholded precision and recall metrics to compare our global SuperPADL controller against two baselines trained on the same dataset of 5587 motions. Our first baseline directly applies PADL on the full dataset. Additionally, we train a PADL+BC controller on the entire dataset, instead of on a group of 20 motions. We use the same language encoder architecture as SuperPADL for both baselines and focus on evaluating the motion quality of each method when training on thousands of motions. The policy network sizes are held constant across all three methods. For the PADL and PADL+BC runs, the critic the discriminator are also appropriately scaled up in size.

We report our thresholded precision and recall metrics in Figure~\ref{fig:global-metrics}
and Table~\ref{tab:global-metrics}, observing that both baselines achieve lower precision and recall scores than SuperPADL. Qualitatively, these baseline networks are unable to do much more than stay upright and stumble around, appearing to respond very little to the user's text command. The PADL+BC network will occasionally reproduce short snippets of simple motions such as jogging. We emphasize that these baselines attempt to apply adversarial reinforcement learning objectives at scale, while SuperPADL only trains small-scale policies with RL. Instead, SuperPADL relies exclusively on supervised learning (through DAGGER) to train the global controller \citep{ross2011reduction}.

We assess the ability of the SuperPADL global controller to transition between skills in Appendix~A. The global controller can successfully transition (i.e. not fall) over 90\% of the time, even when transitioning between two skills from different motion groups. Additionally, we evaluate SuperPADL's ability to respond to language commands in Appendix~B, showing that human raters are able to match animations to the appropriate caption a majority of the time.

\begin{table}[]
    \centering
    \caption{Measuring area-under-curve (AUC) motion quality metrics for different global controller objectives. Using adversarial RL on datasets containing thousands of motions is ineffective, leading to policies that are largely unresponsive to text commands.}
\begin{tabular}{lrr}
\toprule
   \textbf{Method} & \textbf{Precision AUC} & \textbf{Recall AUC} \\
\midrule
SuperPADL &      1.18 &   1.11 \\
  PADL+BC &      1.12 &   0.73 \\
     PADL &      0.99 &   0.70 \\
\bottomrule
\end{tabular}
    \label{tab:global-metrics}
\end{table}

\subsection{Evaluating Group Controllers}
\label{sec:group-results}

\begin{figure}
    \centering
    \subfigure
    {\includegraphics[width=0.95\columnwidth]{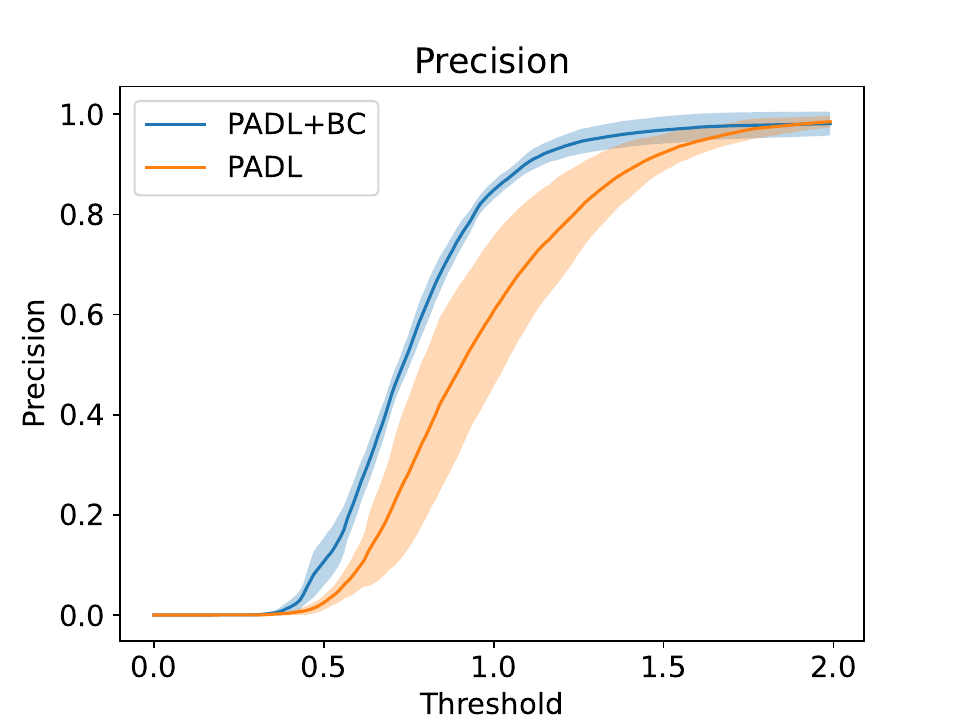}}
    \subfigure
    {\includegraphics[width=0.95\columnwidth]{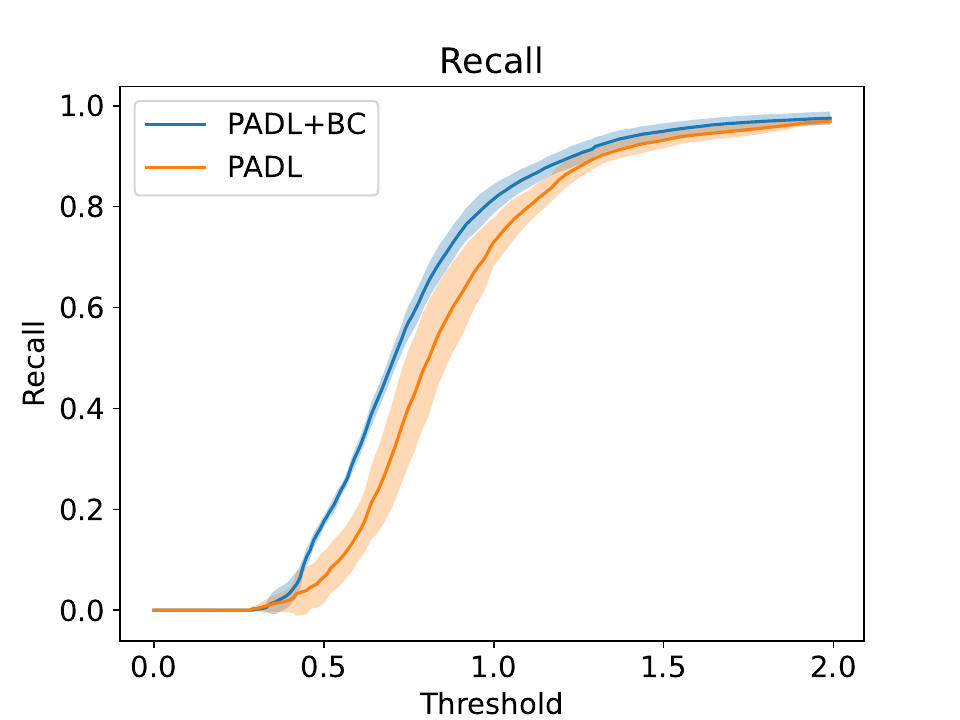}}
    \caption{Thresholded precision and recall metrics for our PADL+BC group controller and a PADL baseline. The PADL+BC controllers record stronger scores on both metrics. Standard deviation is calculated across four trained policies, each trained on a distinct motion group.}
    \label{fig:group-metrics}
\end{figure}

We also assess the motion quality of our PADL+BC group controllers when compared against a pure-PADL baseline.
We randomly select four groups of 20 motions from our dataset and train a controller using both approaches on each group. Note that unlike the PADL models trained in \citet{2022-SA-PADL}, we train pure-PADL group controllers without any language conditioning, instead using the same simple motion index embedding as our PADL+BC controllers. Additionally, we measure the training time required for each network when using a single A40 GPU.

We report our results in Figure~\ref{fig:group-metrics} and Table~\ref{tab:group-metrics}. We observe that group controllers trained with PADL+BC are able to generate higher-quality motions than vanilla PADL controllers while simultaneously requiring much less GPU time. Qualitatively, we observe that PADL+BC models are less prone to imitating only a subsection of a reference motion than PADL policies. Additionally, we find that PADL+BC models seem to be more successful at looping their generations and avoiding getting stuck. Note that the GPU training time reported for PADL+BC models does not include the time required to train the 20 tracking policies that the group controller distills from. A proper accounting of end-to-end compute cost should include these prerequisite training steps as well. However, as shown in Figure~\ref{fig:tracking-times}, the mean tracking policy time is under one hour, therefore even when considering the cost of training tracking experts, the total cost of training one PADL+BC group controller is lower than that of one pure-PADL group controller. Additionally, the training of 20 separate tracking policies can be more easily distributed across multiple GPUs than the training of a single group controller.

\begin{table}[]
    \centering
    \caption{Measuring area-under-curve (AUC) motion quality metrics and policy training time for our work's PADL+BC group controller and a pure-PADL group controller baseline. The supervision signal added to PADL+BC allows it to attain a higher motion quality while training in significantly less time than the baseline. Standard deviation is calculated across four trained policies, each trained on a distinct motion group.}
\begin{tabular}{lrrr}
\toprule
 \textbf{Method} &   \textbf{Precision AUC} &      \textbf{Recall AUC} & \textbf{Training Time}\\
\midrule
PADL+BC & 1.21 ± 0.03 & 1.21 ± 0.02 & 12h \\
   PADL & 1.02 ± 0.11 & 1.10 ± 0.05 & 67h \\
\bottomrule
\end{tabular}
    \label{tab:group-metrics}
\end{table}

\section{Discussion}

In this work we presented SuperPADL, a framework for training physics-based text-conditioned animation models on large datasets. Our approach is predicated on the observations that kinematic motion models, using supervised learning objectives, are able to scale to datasets containing thousands of motions, while RL-based approaches can struggle beyond several hundred motions. In light of this, we employ a progressive distillation process, where we first train small expert policies using RL and then iteratively distill them into larger, more capable networks. Our final controller is able to reproduce skills from a dataset of over 5000 motions and naturally transition in response to changing user commands.

While SuperPADL is able to reproduce many motions with high quality, the network can still struggle with some highly dynamic motions, such as ballet dances or jumps. A limitation of our current approach is that any poorly-reproduced skills (or other flaws) in a group controller will cascade into the global controller during distillation. Additionally, the global controller can still fall over, particularly when asked to transition during a difficult motion. For example, the character will often lose its balance when it is asked to transition mid-kick when it is balancing on one leg with the other leg extended. Altering the timing of motion transitions to avoid these sensitive regions can make the transitions much more reliable. 

We are excited about future work that continues to scale physics-based text-to-motion models to even larger datasets. SuperPADL only uses a fraction of the motions in AMASS and in particular does not train on very long motion capture sequences. While the PADL policies trained in \citet{2022-SA-PADL} were only conditioned on the current character state, SuperPADL is trained on a history of past states. Future architectures that are given an even wider context may be able to learn from longer reference motions. Additionally, we are interested in exploring alternative combinations of RL and supervised learning for physics-based animation. SuperPADL's global controller is a relatively simple deterministic network, and a more sophisticated generative model setup might be better at modelling multi-modal motion distributions. Since the final distillation stage of SuperPADL is purely supervised, it should be possible to train the global controller as a diffusion model using a denoising objective on the target actions. This would give users access to many of the customization techniques that have been successful with text-to-image models, such as guidance and different noise schedulers \citep{ho2022classifierfree}. Overall, we hope that our work contributes to the development of more capable physics-based animation models as well as more powerful, accessible animation tools.

\newpage

\bibliographystyle{ACM-Reference-Format}
\bibliography{main}


\begin{thebibliography}{64}


\ifx \showCODEN    \undefined \def \showCODEN     #1{\unskip}     \fi
\ifx \showDOI      \undefined \def \showDOI       #1{#1}\fi
\ifx \showISBNx    \undefined \def \showISBNx     #1{\unskip}     \fi
\ifx \showISBNxiii \undefined \def \showISBNxiii  #1{\unskip}     \fi
\ifx \showISSN     \undefined \def \showISSN      #1{\unskip}     \fi
\ifx \showLCCN     \undefined \def \showLCCN      #1{\unskip}     \fi
\ifx \shownote     \undefined \def \shownote      #1{#1}          \fi
\ifx \showarticletitle \undefined \def \showarticletitle #1{#1}   \fi
\ifx \showURL      \undefined \def \showURL       {\relax}        \fi
\providecommand\bibfield[2]{#2}
\providecommand\bibinfo[2]{#2}
\providecommand\natexlab[1]{#1}
\providecommand\showeprint[2][]{arXiv:#2}

\bibitem[Agrawal and van~de Panne(2016)]%
        {taskBasedLocomotion2016}
\bibfield{author}{\bibinfo{person}{Shailen Agrawal} {and} \bibinfo{person}{Michiel van~de Panne}.} \bibinfo{year}{2016}\natexlab{}.
\newblock \showarticletitle{Task-based Locomotion}.
\newblock \bibinfo{journal}{\emph{ACM Transactions on Graphics (Proc. SIGGRAPH 2016)}} \bibinfo{volume}{35}, \bibinfo{number}{4} (\bibinfo{year}{2016}).
\newblock


\bibitem[Ahuja and Morency(2019)]%
        {Language2Pose2019}
\bibfield{author}{\bibinfo{person}{C. Ahuja} {and} \bibinfo{person}{L. Morency}.} \bibinfo{year}{2019}\natexlab{}.
\newblock \showarticletitle{Language2Pose: Natural Language Grounded Pose Forecasting}. In \bibinfo{booktitle}{\emph{2019 International Conference on 3D Vision (3DV)}}. \bibinfo{publisher}{IEEE Computer Society}, \bibinfo{address}{Los Alamitos, CA, USA}, \bibinfo{pages}{719--728}.
\newblock
\urldef\tempurl%
\url{https://doi.org/10.1109/3DV.2019.00084}
\showDOI{\tempurl}


\bibitem[Athanasiou et~al\mbox{.}(2022)]%
        {TEACH:3DV:2022}
\bibfield{author}{\bibinfo{person}{Nikos Athanasiou}, \bibinfo{person}{Mathis Petrovich}, \bibinfo{person}{Michael~J. Black}, {and} \bibinfo{person}{G{\"u}l Varol}.} \bibinfo{year}{2022}\natexlab{}.
\newblock \showarticletitle{{TEACH}: {T}emporal {A}ction {C}ompositions for {3D} {H}umans}. In \bibinfo{booktitle}{\emph{{International Conference on 3D Vision (3DV)}}}.
\newblock


\bibitem[Ba et~al\mbox{.}(2016)]%
        {ba2016layer}
\bibfield{author}{\bibinfo{person}{Jimmy~Lei Ba}, \bibinfo{person}{Jamie~Ryan Kiros}, {and} \bibinfo{person}{Geoffrey~E. Hinton}.} \bibinfo{year}{2016}\natexlab{}.
\newblock \bibinfo{title}{Layer Normalization}.
\newblock
\newblock
\showeprint[arxiv]{1607.06450}~[stat.ML]


\bibitem[Bergamin et~al\mbox{.}(2019)]%
        {DreCon2019}
\bibfield{author}{\bibinfo{person}{Kevin Bergamin}, \bibinfo{person}{Simon Clavet}, \bibinfo{person}{Daniel Holden}, {and} \bibinfo{person}{James~Richard Forbes}.} \bibinfo{year}{2019}\natexlab{}.
\newblock \showarticletitle{DReCon: Data-Driven Responsive Control of Physics-Based Characters}.
\newblock \bibinfo{journal}{\emph{ACM Trans. Graph.}} \bibinfo{volume}{38}, \bibinfo{number}{6}, Article \bibinfo{articleno}{206} (\bibinfo{date}{Nov.} \bibinfo{year}{2019}), \bibinfo{numpages}{11}~pages.
\newblock
\showISSN{0730-0301}
\urldef\tempurl%
\url{https://doi.org/10.1145/3355089.3356536}
\showDOI{\tempurl}


\bibitem[Brown et~al\mbox{.}(2020)]%
        {GPT2020}
\bibfield{author}{\bibinfo{person}{Tom~B. Brown}, \bibinfo{person}{Benjamin Mann}, \bibinfo{person}{Nick Ryder}, \bibinfo{person}{Melanie Subbiah}, \bibinfo{person}{Jared Kaplan}, \bibinfo{person}{Prafulla Dhariwal}, \bibinfo{person}{Arvind Neelakantan}, \bibinfo{person}{Pranav Shyam}, \bibinfo{person}{Girish Sastry}, \bibinfo{person}{Amanda Askell}, \bibinfo{person}{Sandhini Agarwal}, \bibinfo{person}{Ariel Herbert{-}Voss}, \bibinfo{person}{Gretchen Krueger}, \bibinfo{person}{Tom Henighan}, \bibinfo{person}{Rewon Child}, \bibinfo{person}{Aditya Ramesh}, \bibinfo{person}{Daniel~M. Ziegler}, \bibinfo{person}{Jeffrey Wu}, \bibinfo{person}{Clemens Winter}, \bibinfo{person}{Christopher Hesse}, \bibinfo{person}{Mark Chen}, \bibinfo{person}{Eric Sigler}, \bibinfo{person}{Mateusz Litwin}, \bibinfo{person}{Scott Gray}, \bibinfo{person}{Benjamin Chess}, \bibinfo{person}{Jack Clark}, \bibinfo{person}{Christopher Berner}, \bibinfo{person}{Sam McCandlish}, \bibinfo{person}{Alec Radford}, \bibinfo{person}{Ilya Sutskever},
  {and} \bibinfo{person}{Dario Amodei}.} \bibinfo{year}{2020}\natexlab{}.
\newblock \showarticletitle{Language Models are Few-Shot Learners}.
\newblock \bibinfo{journal}{\emph{CoRR}}  \bibinfo{volume}{abs/2005.14165} (\bibinfo{year}{2020}).
\newblock
\showeprint[arXiv]{2005.14165}
\urldef\tempurl%
\url{https://arxiv.org/abs/2005.14165}
\showURL{%
\tempurl}


\bibitem[Clegg et~al\mbox{.}(2018)]%
        {DressClegg2018}
\bibfield{author}{\bibinfo{person}{Alexander Clegg}, \bibinfo{person}{Wenhao Yu}, \bibinfo{person}{Jie Tan}, \bibinfo{person}{C.~Karen Liu}, {and} \bibinfo{person}{Greg Turk}.} \bibinfo{year}{2018}\natexlab{}.
\newblock \showarticletitle{Learning to Dress: Synthesizing Human Dressing Motion via Deep Reinforcement Learning}.
\newblock \bibinfo{journal}{\emph{ACM Trans. Graph.}} \bibinfo{volume}{37}, \bibinfo{number}{6}, Article \bibinfo{articleno}{179} (\bibinfo{date}{dec} \bibinfo{year}{2018}), \bibinfo{numpages}{10}~pages.
\newblock
\showISSN{0730-0301}
\urldef\tempurl%
\url{https://doi.org/10.1145/3272127.3275048}
\showDOI{\tempurl}


\bibitem[Coros et~al\mbox{.}(2009)]%
        {Coros09}
\bibfield{author}{\bibinfo{person}{Stelian Coros}, \bibinfo{person}{Philippe Beaudoin}, {and} \bibinfo{person}{Michiel van~de Panne}.} \bibinfo{year}{2009}\natexlab{}.
\newblock \showarticletitle{Robust Task-based Control Policies for Physics-based Characters}.
\newblock \bibinfo{journal}{\emph{ACM Trans. Graph. (Proc. SIGGRAPH Asia)}} \bibinfo{volume}{28}, \bibinfo{number}{5} (\bibinfo{year}{2009}), \bibinfo{pages}{Article 170}.
\newblock


\bibitem[da~Silva et~al\mbox{.}(2008)]%
        {Silva2008SimulationOH}
\bibfield{author}{\bibinfo{person}{Marco da Silva}, \bibinfo{person}{Yeuhi Abe}, {and} \bibinfo{person}{Jovan Popovi{\'c}}.} \bibinfo{year}{2008}\natexlab{}.
\newblock \showarticletitle{Simulation of Human Motion Data using Short‐Horizon Model‐Predictive Control}.
\newblock \bibinfo{journal}{\emph{Computer Graphics Forum}}  \bibinfo{volume}{27} (\bibinfo{year}{2008}).
\newblock


\bibitem[Dabral et~al\mbox{.}(2023)]%
        {dabral2022mofusion}
\bibfield{author}{\bibinfo{person}{Rishabh Dabral}, \bibinfo{person}{Muhammad~Hamza Mughal}, \bibinfo{person}{Vladislav Golyanik}, {and} \bibinfo{person}{Christian Theobalt}.} \bibinfo{year}{2023}\natexlab{}.
\newblock \showarticletitle{MoFusion: A Framework for Denoising-Diffusion-based Motion Synthesis}. In \bibinfo{booktitle}{\emph{Computer Vision and Pattern Recognition (CVPR)}}.
\newblock


\bibitem[de~Lasa et~al\mbox{.}(2010)]%
        {delasa2010}
\bibfield{author}{\bibinfo{person}{Martin de Lasa}, \bibinfo{person}{Igor Mordatch}, {and} \bibinfo{person}{Aaron Hertzmann}.} \bibinfo{year}{2010}\natexlab{}.
\newblock \showarticletitle{Feature-Based Locomotion Controllers}.
\newblock \bibinfo{journal}{\emph{ACM Transactions on Graphics}} \bibinfo{volume}{29}, \bibinfo{number}{3} (\bibinfo{year}{2010}).
\newblock


\bibitem[Devlin et~al\mbox{.}(2018)]%
        {bert2018}
\bibfield{author}{\bibinfo{person}{Jacob Devlin}, \bibinfo{person}{Ming-Wei Chang}, \bibinfo{person}{Kenton Lee}, {and} \bibinfo{person}{Kristina Toutanova}.} \bibinfo{year}{2018}\natexlab{}.
\newblock \bibinfo{title}{BERT: Pre-training of Deep Bidirectional Transformers for Language Understanding}.
\newblock
\newblock
\urldef\tempurl%
\url{https://doi.org/10.48550/ARXIV.1810.04805}
\showDOI{\tempurl}


\bibitem[Guo et~al\mbox{.}(2022)]%
        {HumanML3D2022}
\bibfield{author}{\bibinfo{person}{Chuan Guo}, \bibinfo{person}{Shihao Zou}, \bibinfo{person}{Xinxin Zuo}, \bibinfo{person}{Sen Wang}, \bibinfo{person}{Wei Ji}, \bibinfo{person}{Xingyu Li}, {and} \bibinfo{person}{Li Cheng}.} \bibinfo{year}{2022}\natexlab{}.
\newblock \showarticletitle{Generating Diverse and Natural 3D Human Motions From Text}. In \bibinfo{booktitle}{\emph{Proceedings of the IEEE/CVF Conference on Computer Vision and Pattern Recognition (CVPR)}}. \bibinfo{pages}{5152--5161}.
\newblock


\bibitem[Ho and Salimans(2022)]%
        {ho2022classifierfree}
\bibfield{author}{\bibinfo{person}{Jonathan Ho} {and} \bibinfo{person}{Tim Salimans}.} \bibinfo{year}{2022}\natexlab{}.
\newblock \bibinfo{title}{Classifier-Free Diffusion Guidance}.
\newblock
\newblock
\showeprint[arxiv]{2207.12598}~[cs.LG]


\bibitem[Hodgins et~al\mbox{.}(1995)]%
        {AthleticsHodgins1995}
\bibfield{author}{\bibinfo{person}{Jessica~K. Hodgins}, \bibinfo{person}{Wayne~L. Wooten}, \bibinfo{person}{David~C. Brogan}, {and} \bibinfo{person}{James~F. O'Brien}.} \bibinfo{year}{1995}\natexlab{}.
\newblock \showarticletitle{Animating Human Athletics}. In \bibinfo{booktitle}{\emph{Proceedings of the 22nd Annual Conference on Computer Graphics and Interactive Techniques}} \emph{(\bibinfo{series}{SIGGRAPH '95})}. \bibinfo{publisher}{Association for Computing Machinery}, \bibinfo{address}{New York, NY, USA}, \bibinfo{pages}{71–78}.
\newblock
\showISBNx{0897917014}
\urldef\tempurl%
\url{https://doi.org/10.1145/218380.218414}
\showDOI{\tempurl}


\bibitem[Holden et~al\mbox{.}(2017)]%
        {PFNN2017}
\bibfield{author}{\bibinfo{person}{Daniel Holden}, \bibinfo{person}{Taku Komura}, {and} \bibinfo{person}{Jun Saito}.} \bibinfo{year}{2017}\natexlab{}.
\newblock \showarticletitle{Phase-Functioned Neural Networks for Character Control}.
\newblock \bibinfo{journal}{\emph{ACM Trans. Graph.}} \bibinfo{volume}{36}, \bibinfo{number}{4}, Article \bibinfo{articleno}{42} (\bibinfo{date}{jul} \bibinfo{year}{2017}), \bibinfo{numpages}{13}~pages.
\newblock
\showISSN{0730-0301}
\urldef\tempurl%
\url{https://doi.org/10.1145/3072959.3073663}
\showDOI{\tempurl}


\bibitem[Jiang et~al\mbox{.}(2023)]%
        {jiang2023motiongpt}
\bibfield{author}{\bibinfo{person}{Biao Jiang}, \bibinfo{person}{Xin Chen}, \bibinfo{person}{Wen Liu}, \bibinfo{person}{Jingyi Yu}, \bibinfo{person}{Gang Yu}, {and} \bibinfo{person}{Tao Chen}.} \bibinfo{year}{2023}\natexlab{}.
\newblock \showarticletitle{MotionGPT: Human Motion as a Foreign Language}.
\newblock \bibinfo{journal}{\emph{arXiv preprint arXiv:2306.14795}} (\bibinfo{year}{2023}).
\newblock


\bibitem[Juravsky et~al\mbox{.}(2022)]%
        {2022-SA-PADL}
\bibfield{author}{\bibinfo{person}{Jordan Juravsky}, \bibinfo{person}{Yunrong Guo}, \bibinfo{person}{Sanja Fidler}, {and} \bibinfo{person}{Xue~Bin Peng}.} \bibinfo{year}{2022}\natexlab{}.
\newblock \showarticletitle{PADL: Language-Directed Physics-Based Character Control}. In \bibinfo{booktitle}{\emph{SIGGRAPH Asia 2022 Conference Papers}} (Daegu, Republic of Korea) \emph{(\bibinfo{series}{SA '22})}. \bibinfo{publisher}{Association for Computing Machinery}, \bibinfo{address}{New York, NY, USA}, Article \bibinfo{articleno}{19}, \bibinfo{numpages}{9}~pages.
\newblock
\showISBNx{9781450394703}
\urldef\tempurl%
\url{https://doi.org/10.1145/3550469.3555391}
\showDOI{\tempurl}


\bibitem[Lee et~al\mbox{.}(2021b)]%
        {TimeCritical2021}
\bibfield{author}{\bibinfo{person}{Kyungho Lee}, \bibinfo{person}{Sehee Min}, \bibinfo{person}{Sunmin Lee}, {and} \bibinfo{person}{Jehee Lee}.} \bibinfo{year}{2021}\natexlab{b}.
\newblock \showarticletitle{Learning Time-Critical Responses for Interactive Character Control}.
\newblock \bibinfo{journal}{\emph{ACM Trans. Graph.}} \bibinfo{volume}{40}, \bibinfo{number}{4}, Article \bibinfo{articleno}{147} (\bibinfo{date}{jul} \bibinfo{year}{2021}), \bibinfo{numpages}{11}~pages.
\newblock
\showISSN{0730-0301}
\urldef\tempurl%
\url{https://doi.org/10.1145/3450626.3459826}
\showDOI{\tempurl}


\bibitem[Lee et~al\mbox{.}(2021a)]%
        {Lee2021Parameterized}
\bibfield{author}{\bibinfo{person}{Seyoung Lee}, \bibinfo{person}{Sunmin Lee}, \bibinfo{person}{Yongwoo Lee}, {and} \bibinfo{person}{Jehee Lee}.} \bibinfo{year}{2021}\natexlab{a}.
\newblock \showarticletitle{Learning a family of motor skills from a single motion clip}.
\newblock \bibinfo{journal}{\emph{ACM Trans. Graph.}} \bibinfo{volume}{40}, \bibinfo{number}{4}, Article \bibinfo{articleno}{93} (\bibinfo{year}{2021}).
\newblock


\bibitem[Lee et~al\mbox{.}(2010a)]%
        {DataDrivenLee2010}
\bibfield{author}{\bibinfo{person}{Yoonsang Lee}, \bibinfo{person}{Sungeun Kim}, {and} \bibinfo{person}{Jehee Lee}.} \bibinfo{year}{2010}\natexlab{a}.
\newblock \showarticletitle{Data-Driven Biped Control}.
\newblock \bibinfo{journal}{\emph{ACM Trans. Graph.}} \bibinfo{volume}{29}, \bibinfo{number}{4}, Article \bibinfo{articleno}{129} (\bibinfo{date}{July} \bibinfo{year}{2010}), \bibinfo{numpages}{8}~pages.
\newblock
\showISSN{0730-0301}
\urldef\tempurl%
\url{https://doi.org/10.1145/1778765.1781155}
\showDOI{\tempurl}


\bibitem[Lee et~al\mbox{.}(2010b)]%
        {MotionFields2010}
\bibfield{author}{\bibinfo{person}{Yongjoon Lee}, \bibinfo{person}{Kevin Wampler}, \bibinfo{person}{Gilbert Bernstein}, \bibinfo{person}{Jovan Popovi\'{c}}, {and} \bibinfo{person}{Zoran Popovi\'{c}}.} \bibinfo{year}{2010}\natexlab{b}.
\newblock \showarticletitle{Motion Fields for Interactive Character Locomotion}.
\newblock \bibinfo{journal}{\emph{ACM Trans. Graph.}} \bibinfo{volume}{29}, \bibinfo{number}{6}, Article \bibinfo{articleno}{138} (\bibinfo{date}{dec} \bibinfo{year}{2010}), \bibinfo{numpages}{8}~pages.
\newblock
\showISSN{0730-0301}
\urldef\tempurl%
\url{https://doi.org/10.1145/1882261.1866160}
\showDOI{\tempurl}


\bibitem[Lin et~al\mbox{.}(2018)]%
        {linvigil18}
\bibfield{author}{\bibinfo{person}{Angela~S. Lin}, \bibinfo{person}{Lemeng Wu}, \bibinfo{person}{Rodolfo Corona}, \bibinfo{person}{Kevin Tai}, \bibinfo{person}{Qixing Huang}, {and} \bibinfo{person}{Raymond~J. Mooney}.} \bibinfo{year}{2018}\natexlab{}.
\newblock \showarticletitle{Generating Animated Videos of Human Activities from Natural Language Descriptions}. In \bibinfo{booktitle}{\emph{Proceedings of the Visually Grounded Interaction and Language Workshop at NeurIPS 2018}}.
\newblock
\urldef\tempurl%
\url{http://www.cs.utexas.edu/users/ai-labpub-view.php?PubID=127730}
\showURL{%
\tempurl}


\bibitem[Ling et~al\mbox{.}(2020)]%
        {2020-TOG-MVAE}
\bibfield{author}{\bibinfo{person}{Hung~Yu Ling}, \bibinfo{person}{Fabio Zinno}, \bibinfo{person}{George Cheng}, {and} \bibinfo{person}{Michiel van~de Panne}.} \bibinfo{year}{2020}\natexlab{}.
\newblock \showarticletitle{Character Controllers Using Motion VAEs}.
\newblock \bibinfo{journal}{\emph{ACM Trans. Graph.}} \bibinfo{volume}{39}, \bibinfo{number}{4} (\bibinfo{year}{2020}).
\newblock


\bibitem[Liu and Hodgins(2018)]%
        {BasketballLiu2018}
\bibfield{author}{\bibinfo{person}{Libin Liu} {and} \bibinfo{person}{Jessica Hodgins}.} \bibinfo{year}{2018}\natexlab{}.
\newblock \showarticletitle{Learning basketball dribbling skills using trajectory optimization and deep reinforcement learning}.
\newblock \bibinfo{journal}{\emph{ACM Trans. Graph.}} \bibinfo{volume}{37}, \bibinfo{number}{4}, Article \bibinfo{articleno}{142} (\bibinfo{date}{jul} \bibinfo{year}{2018}), \bibinfo{numpages}{14}~pages.
\newblock
\showISSN{0730-0301}
\urldef\tempurl%
\url{https://doi.org/10.1145/3197517.3201315}
\showDOI{\tempurl}


\bibitem[Liu et~al\mbox{.}(2019)]%
        {robertaLiu}
\bibfield{author}{\bibinfo{person}{Yinhan Liu}, \bibinfo{person}{Myle Ott}, \bibinfo{person}{Naman Goyal}, \bibinfo{person}{Jingfei Du}, \bibinfo{person}{Mandar Joshi}, \bibinfo{person}{Danqi Chen}, \bibinfo{person}{Omer Levy}, \bibinfo{person}{Mike Lewis}, \bibinfo{person}{Luke Zettlemoyer}, {and} \bibinfo{person}{Veselin Stoyanov}.} \bibinfo{year}{2019}\natexlab{}.
\newblock \bibinfo{title}{RoBERTa: A Robustly Optimized BERT Pretraining Approach}.
\newblock
\newblock
\urldef\tempurl%
\url{https://doi.org/10.48550/ARXIV.1907.11692}
\showDOI{\tempurl}


\bibitem[Luo et~al\mbox{.}(2023)]%
        {Luo2023PerpetualHC}
\bibfield{author}{\bibinfo{person}{Zhengyi Luo}, \bibinfo{person}{Jinkun Cao}, \bibinfo{person}{Alexander~W. Winkler}, \bibinfo{person}{Kris Kitani}, {and} \bibinfo{person}{Weipeng Xu}.} \bibinfo{year}{2023}\natexlab{}.
\newblock \showarticletitle{Perpetual Humanoid Control for Real-time Simulated Avatars}. In \bibinfo{booktitle}{\emph{International Conference on Computer Vision (ICCV)}}.
\newblock


\bibitem[Mahmood et~al\mbox{.}(2019)]%
        {mahmood2019amass}
\bibfield{author}{\bibinfo{person}{Naureen Mahmood}, \bibinfo{person}{Nima Ghorbani}, \bibinfo{person}{Nikolaus~F. Troje}, \bibinfo{person}{Gerard Pons-Moll}, {and} \bibinfo{person}{Michael~J. Black}.} \bibinfo{year}{2019}\natexlab{}.
\newblock \bibinfo{title}{AMASS: Archive of Motion Capture as Surface Shapes}.
\newblock
\newblock
\showeprint[arxiv]{1904.03278}~[cs.CV]


\bibitem[Makoviychuk et~al\mbox{.}(2021)]%
        {IsaacGym2021}
\bibfield{author}{\bibinfo{person}{Viktor Makoviychuk}, \bibinfo{person}{Lukasz Wawrzyniak}, \bibinfo{person}{Yunrong Guo}, \bibinfo{person}{Michelle Lu}, \bibinfo{person}{Kier Storey}, \bibinfo{person}{Miles Macklin}, \bibinfo{person}{David Hoeller}, \bibinfo{person}{Nikita Rudin}, \bibinfo{person}{Arthur Allshire}, \bibinfo{person}{Ankur Handa}, {and} \bibinfo{person}{Gavriel State}.} \bibinfo{year}{2021}\natexlab{}.
\newblock \showarticletitle{Isaac Gym: High Performance GPU-Based Physics Simulation For Robot Learning}.
\newblock \bibinfo{journal}{\emph{CoRR}}  \bibinfo{volume}{abs/2108.10470} (\bibinfo{year}{2021}).
\newblock
\showeprint[arXiv]{2108.10470}
\urldef\tempurl%
\url{https://arxiv.org/abs/2108.10470}
\showURL{%
\tempurl}


\bibitem[Merel et~al\mbox{.}(2019)]%
        {merel2018neural}
\bibfield{author}{\bibinfo{person}{Josh Merel}, \bibinfo{person}{Leonard Hasenclever}, \bibinfo{person}{Alexandre Galashov}, \bibinfo{person}{Arun Ahuja}, \bibinfo{person}{Vu Pham}, \bibinfo{person}{Greg Wayne}, \bibinfo{person}{Yee~Whye Teh}, {and} \bibinfo{person}{Nicolas Heess}.} \bibinfo{year}{2019}\natexlab{}.
\newblock \showarticletitle{Neural Probabilistic Motor Primitives for Humanoid Control}. In \bibinfo{booktitle}{\emph{International Conference on Learning Representations}}.
\newblock
\urldef\tempurl%
\url{https://openreview.net/forum?id=BJl6TjRcY7}
\showURL{%
\tempurl}


\bibitem[Merel et~al\mbox{.}(2020)]%
        {CatchCarry2020}
\bibfield{author}{\bibinfo{person}{Josh Merel}, \bibinfo{person}{Saran Tunyasuvunakool}, \bibinfo{person}{Arun Ahuja}, \bibinfo{person}{Yuval Tassa}, \bibinfo{person}{Leonard Hasenclever}, \bibinfo{person}{Vu Pham}, \bibinfo{person}{Tom Erez}, \bibinfo{person}{Greg Wayne}, {and} \bibinfo{person}{Nicolas Heess}.} \bibinfo{year}{2020}\natexlab{}.
\newblock \showarticletitle{Catch \& Carry: reusable neural controllers for vision-guided whole-body tasks}.
\newblock \bibinfo{journal}{\emph{ACM Trans. Graph.}} \bibinfo{volume}{39}, \bibinfo{number}{4}, Article \bibinfo{articleno}{39} (\bibinfo{date}{aug} \bibinfo{year}{2020}), \bibinfo{numpages}{14}~pages.
\newblock
\showISSN{0730-0301}
\urldef\tempurl%
\url{https://doi.org/10.1145/3386569.3392474}
\showDOI{\tempurl}


\bibitem[Peng et~al\mbox{.}(2018)]%
        {2018-TOG-deepMimic}
\bibfield{author}{\bibinfo{person}{Xue~Bin Peng}, \bibinfo{person}{Pieter Abbeel}, \bibinfo{person}{Sergey Levine}, {and} \bibinfo{person}{Michiel van~de Panne}.} \bibinfo{year}{2018}\natexlab{}.
\newblock \showarticletitle{DeepMimic: Example-guided Deep Reinforcement Learning of Physics-based Character Skills}.
\newblock \bibinfo{journal}{\emph{ACM Trans. Graph.}} \bibinfo{volume}{37}, \bibinfo{number}{4}, Article \bibinfo{articleno}{143} (\bibinfo{date}{July} \bibinfo{year}{2018}), \bibinfo{numpages}{14}~pages.
\newblock
\showISSN{0730-0301}
\urldef\tempurl%
\url{https://doi.org/10.1145/3197517.3201311}
\showDOI{\tempurl}


\bibitem[Peng et~al\mbox{.}(2017)]%
        {2017-TOG-deepLoco}
\bibfield{author}{\bibinfo{person}{Xue~Bin Peng}, \bibinfo{person}{Glen Berseth}, \bibinfo{person}{Kangkang Yin}, {and} \bibinfo{person}{Michiel Van De~Panne}.} \bibinfo{year}{2017}\natexlab{}.
\newblock \showarticletitle{DeepLoco: Dynamic Locomotion Skills Using Hierarchical Deep Reinforcement Learning}.
\newblock \bibinfo{journal}{\emph{ACM Trans. Graph.}} \bibinfo{volume}{36}, \bibinfo{number}{4}, Article \bibinfo{articleno}{41} (\bibinfo{date}{July} \bibinfo{year}{2017}), \bibinfo{numpages}{13}~pages.
\newblock
\showISSN{0730-0301}
\urldef\tempurl%
\url{https://doi.org/10.1145/3072959.3073602}
\showDOI{\tempurl}


\bibitem[Peng et~al\mbox{.}(2022)]%
        {2022-TOG-ASE}
\bibfield{author}{\bibinfo{person}{Xue~Bin Peng}, \bibinfo{person}{Yunrong Guo}, \bibinfo{person}{Lina Halper}, \bibinfo{person}{Sergey Levine}, {and} \bibinfo{person}{Sanja Fidler}.} \bibinfo{year}{2022}\natexlab{}.
\newblock \showarticletitle{ASE: Large-scale Reusable Adversarial Skill Embeddings for Physically Simulated Characters}.
\newblock \bibinfo{journal}{\emph{ACM Trans. Graph.}} \bibinfo{volume}{41}, \bibinfo{number}{4}, Article \bibinfo{articleno}{94} (\bibinfo{date}{July} \bibinfo{year}{2022}).
\newblock


\bibitem[Peng et~al\mbox{.}(2021)]%
        {2021-TOG-AMP}
\bibfield{author}{\bibinfo{person}{Xue~Bin Peng}, \bibinfo{person}{Ze Ma}, \bibinfo{person}{Pieter Abbeel}, \bibinfo{person}{Sergey Levine}, {and} \bibinfo{person}{Angjoo Kanazawa}.} \bibinfo{year}{2021}\natexlab{}.
\newblock \showarticletitle{AMP: Adversarial Motion Priors for Stylized Physics-Based Character Control}.
\newblock \bibinfo{journal}{\emph{ACM Trans. Graph.}} \bibinfo{volume}{40}, \bibinfo{number}{4}, Article \bibinfo{articleno}{1} (\bibinfo{date}{July} \bibinfo{year}{2021}), \bibinfo{numpages}{15}~pages.
\newblock
\urldef\tempurl%
\url{https://doi.org/10.1145/3450626.3459670}
\showDOI{\tempurl}


\bibitem[Petrovich et~al\mbox{.}(2022)]%
        {petrovich22temos}
\bibfield{author}{\bibinfo{person}{Mathis Petrovich}, \bibinfo{person}{Michael~J. Black}, {and} \bibinfo{person}{G{\"u}l Varol}.} \bibinfo{year}{2022}\natexlab{}.
\newblock \showarticletitle{{TEMOS}: Generating diverse human motions from textual descriptions}. In \bibinfo{booktitle}{\emph{European Conference on Computer Vision ({ECCV})}}.
\newblock


\bibitem[Plappert et~al\mbox{.}(2017)]%
        {PlappertMA17}
\bibfield{author}{\bibinfo{person}{Matthias Plappert}, \bibinfo{person}{Christian Mandery}, {and} \bibinfo{person}{Tamim Asfour}.} \bibinfo{year}{2017}\natexlab{}.
\newblock \showarticletitle{Learning a bidirectional mapping between human whole-body motion and natural language using deep recurrent neural networks}.
\newblock \bibinfo{journal}{\emph{CoRR}}  \bibinfo{volume}{abs/1705.06400} (\bibinfo{year}{2017}).
\newblock
\showeprint[arXiv]{1705.06400}
\urldef\tempurl%
\url{http://arxiv.org/abs/1705.06400}
\showURL{%
\tempurl}


\bibitem[Poole et~al\mbox{.}(2023)]%
        {poole2023dreamfusion}
\bibfield{author}{\bibinfo{person}{Ben Poole}, \bibinfo{person}{Ajay Jain}, \bibinfo{person}{Jonathan~T. Barron}, {and} \bibinfo{person}{Ben Mildenhall}.} \bibinfo{year}{2023}\natexlab{}.
\newblock \showarticletitle{DreamFusion: Text-to-3D using 2D Diffusion}. In \bibinfo{booktitle}{\emph{The Eleventh International Conference on Learning Representations}}.
\newblock
\urldef\tempurl%
\url{https://openreview.net/forum?id=FjNys5c7VyY}
\showURL{%
\tempurl}


\bibitem[Punnakkal et~al\mbox{.}(2021)]%
        {BABEL:CVPR:2021}
\bibfield{author}{\bibinfo{person}{Abhinanda~R. Punnakkal}, \bibinfo{person}{Arjun Chandrasekaran}, \bibinfo{person}{Nikos Athanasiou}, \bibinfo{person}{Alejandra Quiros-Ramirez}, {and} \bibinfo{person}{Michael~J. Black}.} \bibinfo{year}{2021}\natexlab{}.
\newblock \showarticletitle{{BABEL}: Bodies, Action and Behavior with English Labels}. In \bibinfo{booktitle}{\emph{Proceedings IEEE/CVF Conf.~on Computer Vision and Pattern Recognition (CVPR)}}. \bibinfo{pages}{722--731}.
\newblock


\bibitem[Raffel et~al\mbox{.}(2019)]%
        {t5Raffel}
\bibfield{author}{\bibinfo{person}{Colin Raffel}, \bibinfo{person}{Noam Shazeer}, \bibinfo{person}{Adam Roberts}, \bibinfo{person}{Katherine Lee}, \bibinfo{person}{Sharan Narang}, \bibinfo{person}{Michael Matena}, \bibinfo{person}{Yanqi Zhou}, \bibinfo{person}{Wei Li}, {and} \bibinfo{person}{Peter~J. Liu}.} \bibinfo{year}{2019}\natexlab{}.
\newblock \bibinfo{title}{Exploring the Limits of Transfer Learning with a Unified Text-to-Text Transformer}.
\newblock
\newblock
\urldef\tempurl%
\url{https://doi.org/10.48550/ARXIV.1910.10683}
\showDOI{\tempurl}


\bibitem[Ren et~al\mbox{.}(2023)]%
        {ren2023insactor}
\bibfield{author}{\bibinfo{person}{Jiawei Ren}, \bibinfo{person}{Mingyuan Zhang}, \bibinfo{person}{Cunjun Yu}, \bibinfo{person}{Xiao Ma}, \bibinfo{person}{Liang Pan}, {and} \bibinfo{person}{Ziwei Liu}.} \bibinfo{year}{2023}\natexlab{}.
\newblock \showarticletitle{InsActor: Instruction-driven Physics-based Characters}.
\newblock \bibinfo{journal}{\emph{NeurIPS}} (\bibinfo{year}{2023}).
\newblock


\bibitem[Ross et~al\mbox{.}(2011)]%
        {ross2011reduction}
\bibfield{author}{\bibinfo{person}{Stephane Ross}, \bibinfo{person}{Geoffrey~J. Gordon}, {and} \bibinfo{person}{J.~Andrew Bagnell}.} \bibinfo{year}{2011}\natexlab{}.
\newblock \bibinfo{title}{A Reduction of Imitation Learning and Structured Prediction to No-Regret Online Learning}.
\newblock
\newblock
\showeprint[arxiv]{1011.0686}~[cs.LG]


\bibitem[Saharia et~al\mbox{.}(2022)]%
        {saharia2022photorealistic}
\bibfield{author}{\bibinfo{person}{Chitwan Saharia}, \bibinfo{person}{William Chan}, \bibinfo{person}{Saurabh Saxena}, \bibinfo{person}{Lala Li}, \bibinfo{person}{Jay Whang}, \bibinfo{person}{Emily Denton}, \bibinfo{person}{Seyed Kamyar~Seyed Ghasemipour}, \bibinfo{person}{Raphael Gontijo-Lopes}, \bibinfo{person}{Burcu~Karagol Ayan}, \bibinfo{person}{Tim Salimans}, \bibinfo{person}{Jonathan Ho}, \bibinfo{person}{David~J. Fleet}, {and} \bibinfo{person}{Mohammad Norouzi}.} \bibinfo{year}{2022}\natexlab{}.
\newblock \showarticletitle{Photorealistic Text-to-Image Diffusion Models with Deep Language Understanding}. In \bibinfo{booktitle}{\emph{Advances in Neural Information Processing Systems}}, \bibfield{editor}{\bibinfo{person}{Alice~H. Oh}, \bibinfo{person}{Alekh Agarwal}, \bibinfo{person}{Danielle Belgrave}, {and} \bibinfo{person}{Kyunghyun Cho}} (Eds.).
\newblock
\urldef\tempurl%
\url{https://openreview.net/forum?id=08Yk-n5l2Al}
\showURL{%
\tempurl}


\bibitem[Schulman et~al\mbox{.}(2017)]%
        {PPO2017}
\bibfield{author}{\bibinfo{person}{John Schulman}, \bibinfo{person}{Filip Wolski}, \bibinfo{person}{Prafulla Dhariwal}, \bibinfo{person}{Alec Radford}, {and} \bibinfo{person}{Oleg Klimov}.} \bibinfo{year}{2017}\natexlab{}.
\newblock \showarticletitle{Proximal Policy Optimization Algorithms}.
\newblock \bibinfo{journal}{\emph{CoRR}}  \bibinfo{volume}{abs/1707.06347} (\bibinfo{year}{2017}).
\newblock
\showeprint[arxiv]{1707.06347}
\urldef\tempurl%
\url{http://arxiv.org/abs/1707.06347}
\showURL{%
\tempurl}


\bibitem[Starke et~al\mbox{.}(2019)]%
        {NSM2019}
\bibfield{author}{\bibinfo{person}{Sebastian Starke}, \bibinfo{person}{He Zhang}, \bibinfo{person}{Taku Komura}, {and} \bibinfo{person}{Jun Saito}.} \bibinfo{year}{2019}\natexlab{}.
\newblock \showarticletitle{Neural State Machine for Character-Scene Interactions}.
\newblock \bibinfo{journal}{\emph{ACM Trans. Graph.}} \bibinfo{volume}{38}, \bibinfo{number}{6}, Article \bibinfo{articleno}{209} (\bibinfo{date}{nov} \bibinfo{year}{2019}), \bibinfo{numpages}{14}~pages.
\newblock
\showISSN{0730-0301}
\urldef\tempurl%
\url{https://doi.org/10.1145/3355089.3356505}
\showDOI{\tempurl}


\bibitem[Sun et~al\mbox{.}(2023)]%
        {sun2023prompt}
\bibfield{author}{\bibinfo{person}{Jingkai Sun}, \bibinfo{person}{Qiang Zhang}, \bibinfo{person}{Yiqun Duan}, \bibinfo{person}{Xiaoyang Jiang}, \bibinfo{person}{Chong Cheng}, {and} \bibinfo{person}{Renjing Xu}.} \bibinfo{year}{2023}\natexlab{}.
\newblock \bibinfo{title}{Prompt, Plan, Perform: LLM-based Humanoid Control via Quantized Imitation Learning}.
\newblock
\newblock
\showeprint[arxiv]{2309.11359}~[cs.RO]


\bibitem[Tan et~al\mbox{.}(2014)]%
        {BicycleTan2014}
\bibfield{author}{\bibinfo{person}{Jie Tan}, \bibinfo{person}{Yuting Gu}, \bibinfo{person}{C.~Karen Liu}, {and} \bibinfo{person}{Greg Turk}.} \bibinfo{year}{2014}\natexlab{}.
\newblock \showarticletitle{Learning Bicycle Stunts}.
\newblock \bibinfo{journal}{\emph{ACM Trans. Graph.}} \bibinfo{volume}{33}, \bibinfo{number}{4}, Article \bibinfo{articleno}{50} (\bibinfo{date}{July} \bibinfo{year}{2014}), \bibinfo{numpages}{12}~pages.
\newblock
\showISSN{0730-0301}
\urldef\tempurl%
\url{https://doi.org/10.1145/2601097.2601121}
\showDOI{\tempurl}


\bibitem[Tevet et~al\mbox{.}(2022)]%
        {MotionClipTevet2022}
\bibfield{author}{\bibinfo{person}{Guy Tevet}, \bibinfo{person}{Brian Gordon}, \bibinfo{person}{Amir Hertz}, \bibinfo{person}{Amit~H. Bermano}, {and} \bibinfo{person}{Daniel Cohen-Or}.} \bibinfo{year}{2022}\natexlab{}.
\newblock \bibinfo{title}{MotionCLIP: Exposing Human Motion Generation to CLIP Space}.
\newblock
\newblock
\urldef\tempurl%
\url{https://doi.org/10.48550/ARXIV.2203.08063}
\showDOI{\tempurl}


\bibitem[Tevet et~al\mbox{.}(2023)]%
        {tevet2023human}
\bibfield{author}{\bibinfo{person}{Guy Tevet}, \bibinfo{person}{Sigal Raab}, \bibinfo{person}{Brian Gordon}, \bibinfo{person}{Yoni Shafir}, \bibinfo{person}{Daniel Cohen-or}, {and} \bibinfo{person}{Amit~Haim Bermano}.} \bibinfo{year}{2023}\natexlab{}.
\newblock \showarticletitle{Human Motion Diffusion Model}. In \bibinfo{booktitle}{\emph{The Eleventh International Conference on Learning Representations}}.
\newblock
\urldef\tempurl%
\url{https://openreview.net/forum?id=SJ1kSyO2jwu}
\showURL{%
\tempurl}


\bibitem[Treuille et~al\mbox{.}(2007)]%
        {Treuille2007}
\bibfield{author}{\bibinfo{person}{Adrien Treuille}, \bibinfo{person}{Yongjoon Lee}, {and} \bibinfo{person}{Zoran Popovi\'{c}}.} \bibinfo{year}{2007}\natexlab{}.
\newblock \showarticletitle{Near-Optimal Character Animation with Continuous Control}. In \bibinfo{booktitle}{\emph{ACM SIGGRAPH 2007 Papers}} (San Diego, California) \emph{(\bibinfo{series}{SIGGRAPH '07})}. \bibinfo{publisher}{Association for Computing Machinery}, \bibinfo{address}{New York, NY, USA}, \bibinfo{pages}{7–es}.
\newblock
\showISBNx{9781450378369}
\urldef\tempurl%
\url{https://doi.org/10.1145/1275808.1276386}
\showDOI{\tempurl}


\bibitem[Wagener et~al\mbox{.}(2023)]%
        {wagener2023mocapact}
\bibfield{author}{\bibinfo{person}{Nolan Wagener}, \bibinfo{person}{Andrey Kolobov}, \bibinfo{person}{Felipe~Vieira Frujeri}, \bibinfo{person}{Ricky Loynd}, \bibinfo{person}{Ching-An Cheng}, {and} \bibinfo{person}{Matthew Hausknecht}.} \bibinfo{year}{2023}\natexlab{}.
\newblock \bibinfo{title}{MoCapAct: A Multi-Task Dataset for Simulated Humanoid Control}.
\newblock
\newblock
\showeprint[arxiv]{2208.07363}~[cs.RO]


\bibitem[Wang et~al\mbox{.}(2009)]%
        {BipedWang2009}
\bibfield{author}{\bibinfo{person}{Jack~M. Wang}, \bibinfo{person}{David~J. Fleet}, {and} \bibinfo{person}{Aaron Hertzmann}.} \bibinfo{year}{2009}\natexlab{}.
\newblock \showarticletitle{Optimizing Walking Controllers}. In \bibinfo{booktitle}{\emph{ACM SIGGRAPH Asia 2009 Papers}} (Yokohama, Japan) \emph{(\bibinfo{series}{SIGGRAPH Asia '09})}. \bibinfo{publisher}{Association for Computing Machinery}, \bibinfo{address}{New York, NY, USA}, Article \bibinfo{articleno}{168}, \bibinfo{numpages}{8}~pages.
\newblock
\showISBNx{9781605588582}
\urldef\tempurl%
\url{https://doi.org/10.1145/1661412.1618514}
\showDOI{\tempurl}


\bibitem[Wang et~al\mbox{.}(2012)]%
        {MuscleWang2012}
\bibfield{author}{\bibinfo{person}{Jack~M. Wang}, \bibinfo{person}{Samuel~R. Hamner}, \bibinfo{person}{Scott~L. Delp}, {and} \bibinfo{person}{Vladlen Koltun}.} \bibinfo{year}{2012}\natexlab{}.
\newblock \showarticletitle{Optimizing Locomotion Controllers Using Biologically-Based Actuators and Objectives}.
\newblock \bibinfo{journal}{\emph{ACM Trans. Graph.}} \bibinfo{volume}{31}, \bibinfo{number}{4}, Article \bibinfo{articleno}{25} (\bibinfo{date}{jul} \bibinfo{year}{2012}), \bibinfo{numpages}{11}~pages.
\newblock
\showISSN{0730-0301}
\urldef\tempurl%
\url{https://doi.org/10.1145/2185520.2185521}
\showDOI{\tempurl}


\bibitem[Wang et~al\mbox{.}(2020)]%
        {wang2020unicon}
\bibfield{author}{\bibinfo{person}{Tingwu Wang}, \bibinfo{person}{Yunrong Guo}, \bibinfo{person}{Maria Shugrina}, {and} \bibinfo{person}{Sanja Fidler}.} \bibinfo{year}{2020}\natexlab{}.
\newblock \bibinfo{title}{UniCon: Universal Neural Controller For Physics-based Character Motion}.
\newblock
\newblock
\showeprint[arxiv]{2011.15119}~[cs.GR]


\bibitem[Winkler et~al\mbox{.}(2022)]%
        {QuestSim2022}
\bibfield{author}{\bibinfo{person}{Alexander Winkler}, \bibinfo{person}{Jungdam Won}, {and} \bibinfo{person}{Yuting Ye}.} \bibinfo{year}{2022}\natexlab{}.
\newblock \showarticletitle{QuestSim: Human Motion Tracking from Sparse Sensors with Simulated Avatars}. In \bibinfo{booktitle}{\emph{SIGGRAPH Asia 2022 Conference Papers}} (<conf-loc>, <city>Daegu</city>, <country>Republic of Korea</country>, </conf-loc>) \emph{(\bibinfo{series}{SA '22})}. \bibinfo{publisher}{Association for Computing Machinery}, \bibinfo{address}{New York, NY, USA}, Article \bibinfo{articleno}{2}, \bibinfo{numpages}{8}~pages.
\newblock
\showISBNx{9781450394703}
\urldef\tempurl%
\url{https://doi.org/10.1145/3550469.3555411}
\showDOI{\tempurl}


\bibitem[Won et~al\mbox{.}(2020)]%
        {ScalableWon2020}
\bibfield{author}{\bibinfo{person}{Jungdam Won}, \bibinfo{person}{Deepak Gopinath}, {and} \bibinfo{person}{Jessica Hodgins}.} \bibinfo{year}{2020}\natexlab{}.
\newblock \showarticletitle{A Scalable Approach to Control Diverse Behaviors for Physically Simulated Characters}.
\newblock \bibinfo{journal}{\emph{ACM Trans. Graph.}} \bibinfo{volume}{39}, \bibinfo{number}{4}, Article \bibinfo{articleno}{33} (\bibinfo{date}{jul} \bibinfo{year}{2020}), \bibinfo{numpages}{12}~pages.
\newblock
\showISSN{0730-0301}
\urldef\tempurl%
\url{https://doi.org/10.1145/3386569.3392381}
\showDOI{\tempurl}


\bibitem[Won et~al\mbox{.}(2021)]%
        {2PlayerWon2021}
\bibfield{author}{\bibinfo{person}{Jungdam Won}, \bibinfo{person}{Deepak Gopinath}, {and} \bibinfo{person}{Jessica Hodgins}.} \bibinfo{year}{2021}\natexlab{}.
\newblock \showarticletitle{Control strategies for physically simulated characters performing two-player competitive sports}.
\newblock \bibinfo{journal}{\emph{ACM Trans. Graph.}} \bibinfo{volume}{40}, \bibinfo{number}{4}, Article \bibinfo{articleno}{146} (\bibinfo{date}{jul} \bibinfo{year}{2021}), \bibinfo{numpages}{11}~pages.
\newblock
\showISSN{0730-0301}
\urldef\tempurl%
\url{https://doi.org/10.1145/3450626.3459761}
\showDOI{\tempurl}


\bibitem[Won et~al\mbox{.}(2022)]%
        {PhysVAEWon2022}
\bibfield{author}{\bibinfo{person}{Jungdam Won}, \bibinfo{person}{Deepak Gopinath}, {and} \bibinfo{person}{Jessica Hodgins}.} \bibinfo{year}{2022}\natexlab{}.
\newblock \showarticletitle{Physics-based character controllers using conditional VAEs}.
\newblock \bibinfo{journal}{\emph{ACM Trans. Graph.}} \bibinfo{volume}{41}, \bibinfo{number}{4}, Article \bibinfo{articleno}{96} (\bibinfo{date}{jul} \bibinfo{year}{2022}), \bibinfo{numpages}{12}~pages.
\newblock
\showISSN{0730-0301}
\urldef\tempurl%
\url{https://doi.org/10.1145/3528223.3530067}
\showDOI{\tempurl}


\bibitem[Xie et~al\mbox{.}(2020)]%
        {2020-ALLSTEPS}
\bibfield{author}{\bibinfo{person}{Zhaoming Xie}, \bibinfo{person}{Hung~Yu Ling}, \bibinfo{person}{Nam~Hee Kim}, {and} \bibinfo{person}{Michiel van~de Panne}.} \bibinfo{year}{2020}\natexlab{}.
\newblock \showarticletitle{ALLSTEPS: Curriculum-driven Learning of Stepping Stone Skills}. In \bibinfo{booktitle}{\emph{Proc. ACM SIGGRAPH / Eurographics Symposium on Computer Animation}}.
\newblock


\bibitem[Xie et~al\mbox{.}(2022)]%
        {2022-Soccer-Juggle}
\bibfield{author}{\bibinfo{person}{Zhaoming Xie}, \bibinfo{person}{Sebastian Starke}, \bibinfo{person}{Hung~Yu Ling}, {and} \bibinfo{person}{Michiel van~de Panne}.} \bibinfo{year}{2022}\natexlab{}.
\newblock \showarticletitle{Learning Soccer Juggling Skills with Layer-wise Mixture-of-Experts}.
\newblock  (\bibinfo{year}{2022}).
\newblock


\bibitem[Yao et~al\mbox{.}(2022)]%
        {ControlVAEYao2022}
\bibfield{author}{\bibinfo{person}{Heyuan Yao}, \bibinfo{person}{Zhenhua Song}, \bibinfo{person}{Baoquan Chen}, {and} \bibinfo{person}{Libin Liu}.} \bibinfo{year}{2022}\natexlab{}.
\newblock \showarticletitle{ControlVAE: Model-Based Learning of Generative Controllers for Physics-Based Characters}.
\newblock \bibinfo{journal}{\emph{ACM Trans. Graph.}} \bibinfo{volume}{41}, \bibinfo{number}{6}, Article \bibinfo{articleno}{183} (\bibinfo{date}{nov} \bibinfo{year}{2022}), \bibinfo{numpages}{16}~pages.
\newblock
\showISSN{0730-0301}
\urldef\tempurl%
\url{https://doi.org/10.1145/3550454.3555434}
\showDOI{\tempurl}


\bibitem[Zhang et~al\mbox{.}(2023b)]%
        {zhang2023generating}
\bibfield{author}{\bibinfo{person}{Jianrong Zhang}, \bibinfo{person}{Yangsong Zhang}, \bibinfo{person}{Xiaodong Cun}, \bibinfo{person}{Shaoli Huang}, \bibinfo{person}{Yong Zhang}, \bibinfo{person}{Hongwei Zhao}, \bibinfo{person}{Hongtao Lu}, {and} \bibinfo{person}{Xi Shen}.} \bibinfo{year}{2023}\natexlab{b}.
\newblock \showarticletitle{T2M-GPT: Generating Human Motion from Textual Descriptions with Discrete Representations}. In \bibinfo{booktitle}{\emph{Proceedings of the IEEE/CVF Conference on Computer Vision and Pattern Recognition (CVPR)}}.
\newblock


\bibitem[Zhang et~al\mbox{.}(2023a)]%
        {zhang2023motiongpt}
\bibfield{author}{\bibinfo{person}{Yaqi Zhang}, \bibinfo{person}{Di Huang}, \bibinfo{person}{Bin Liu}, \bibinfo{person}{Shixiang Tang}, \bibinfo{person}{Yan Lu}, \bibinfo{person}{Lu Chen}, \bibinfo{person}{Lei Bai}, \bibinfo{person}{Qi Chu}, \bibinfo{person}{Nenghai Yu}, {and} \bibinfo{person}{Wanli Ouyang}.} \bibinfo{year}{2023}\natexlab{a}.
\newblock \showarticletitle{MotionGPT: Finetuned LLMs are General-Purpose Motion Generators}.
\newblock \bibinfo{journal}{\emph{arXiv preprint arXiv:2306.10900}} (\bibinfo{year}{2023}).
\newblock


\bibitem[Zhang et~al\mbox{.}(2020)]%
        {AmorphousZhang2020}
\bibfield{author}{\bibinfo{person}{Yunbo Zhang}, \bibinfo{person}{Wenhao Yu}, \bibinfo{person}{C.~Karen Liu}, \bibinfo{person}{Charlie Kemp}, {and} \bibinfo{person}{Greg Turk}.} \bibinfo{year}{2020}\natexlab{}.
\newblock \showarticletitle{Learning to Manipulate Amorphous Materials}.
\newblock \bibinfo{journal}{\emph{ACM Trans. Graph.}} \bibinfo{volume}{39}, \bibinfo{number}{6}, Article \bibinfo{articleno}{189} (\bibinfo{date}{nov} \bibinfo{year}{2020}), \bibinfo{numpages}{11}~pages.
\newblock
\showISSN{0730-0301}
\urldef\tempurl%
\url{https://doi.org/10.1145/3414685.3417868}
\showDOI{\tempurl}


\end{thebibliography}

\newpage

\newpage
\clearpage

\appendix
\section{Evaluating Global Controller Transitions}
\label{app:transition}

We evaluate the ability of the SuperPADL global policy to transition between skills without falling over. We assess the policy by rolling out ``transition trajectories'' where the model is initially conditioned on one caption before the caption is changed to that of a different motion. We sample 4096 trajectories, each 10 seconds long with a transition point sampled uniformly between 3 and 7 seconds into the trajectory. In each trajectory, the character is initialized in a default standing position. We classify a transition as successful if the character does not fall over at any point in the trajectory.

We conduct two variations of this experiment. In the first version, we only sample pairs of captions that correspond to two motions from the same motion group (defined in Section~\ref{sec:group}). In the second version, the caption pair must come from motions in two different groups. By evaluating these cases separately, we can assess whether the global controller has learned general transition skills or whether it is limited by the group controllers it is distilled from (which are each only trained to transition between a small group of motions). Our results are summarized in Table~\ref{tab:falls}. We see that the global controller succeeds at the transition over 90\% of the time, regardless of whether caption pairs are sampled from the same group or not.

\begin{table*}
    \centering
    \caption{Evaluating the fraction of successful skill transitions with SuperPADL (i.e. the fraction of transitions where the character does not fall over). }
    \begin{tabular}{|l|l|l|l|}
    \hline
    \textbf{Transition Type} & \textbf{Successful (No Falls)} & \textbf{Fell Before Transition Point} & \textbf{Fell After Transition Point} \\
    \hline
    Caption-Pair from Same Group & 92.70\% & 3.00\% & 4.30\% \\
    \hline
    Caption-Pair from Different Groups & 90.92\% & 3.12\% & 5.96\% \\
    \hline
    \end{tabular}
    \label{tab:falls}
\end{table*}

\section{Evaluating Response to Language Commands}
\label{app:language}

We assess the faithfulness of SuperPADL's generated motions with respect to language commands using human evaluation\footnote{Automated metrics such as FID and R-precision are difficult to apply to our method since they rely on motion encoders that use SMPL character models as input, which are distinct from our physics-based character.}. We generate evaluation questions by sampling 100 captions from our training dataset and recording a 24-second animation from the global controller for each caption. In each animation, we initialize the character in a default standing pose.

We present three human raters with a rendered video of each animation and ask them to select the most appropriate caption from four options, where one answer is the correct caption and the others are randomly sampled from other motions in the dataset. We also present raters with options for ``Nothing applies'' and ``Multiple options apply''. The latter option is helpful to identify cases where the ``incorrect'' alternative captions come from similar motions, for example when both the true caption and an alternative option come from walking clips.

The evaluation results summarize our results in Table~\ref{tab:lang}, with raters discerning the correct caption from SuperPADL's motion a majority of  of the time. When examining looking at the evaluation results, we observed that many ``Nothing applies'' responses correspond to motions where the character struggles to leave the initial standing pose and instead remains 
mostly idle.

\begin{table*}
    \centering
    \caption{Evaluating the ability of human raters to identify the caption that SuperPADL was conditioned on when given four possible options. }
    \begin{tabular}{|l|l|}
    \hline
    \textbf{User Response} & \textbf{Average Selection Frequency} \\
    \hline
    Correct Caption & 57.33\% \\
    \hline
    Incorrect Caption & 19.33\% \\
    \hline
    Multiple Applicable Options & 5.00\% \\
    \hline
    No Applicable Options & 18.33\% \\
    \hline
    \end{tabular}
    \label{tab:lang}
\end{table*}

\section{Architecture and Training Details}
\label{app:arch-train-details}

\subsection{Physics Environment}

All of our experiments are run using the NVIDIA Isaac Gym simulator \citep{IsaacGym2021}. Our character model and observation format matches that of \citet{2022-SA-PADL}, except without a sword and shield. Our action space is 36-dimensional.

\subsection{Expert Tracking Policies}

Each expert tracking policy (and corresponding critic network) is an MLP with two hidden layers containing 1024 and 512 units, respectively. We use an ELU activation function. We encode the reference phase $\phi \in [0,1]$ using two scalars storing $[\sin(\phi), \cos(\phi)]$. Each network is trained using a single A40 GPU.

\subsection{Group Controllers}

When training group controllers, the actors, critics, and discriminators are separate MLP networks with a ReLU activation function and three hidden layers containing [1024, 1024, 512] units. For the actor and critic networks, we provide five frames of character states as input by maintaining a buffer of the 40 most recent frames and sampling every eighth. For the discriminator, we follow the convention of \citet{2022-TOG-ASE} and use the 10 most recent frames as observations. We encode the motion index using an 128-dimensional embedding table. Each network is trained using a single A40 GPU. We train PADL+BC policies for a total of 10000 epochs (including the 2000-epoch warmup period where we only apply the behaviour cloning loss). When training PADL group controller baselines, we train the policy for 54000 epochs, corresponding to approximately 7B frames. 

\subsection{Global Controllers}

All networks used when training global controllers are MLPs with hidden layers containing [3072, 3072, 3072, 2048] units. We provide the policies (and, when applicable, critic networks) with five frames of history using the same method as the group controllers. We train controllers and baselines using eight A40 GPUs.

We train the SuperPADL global controller for 7900 epochs, corresponding to approximately 380K optimization steps, 6B frames of RL training (i.e. 6B online collected samples), and 12 hours of training. Like when training PADL+BC group controllers, we begin with a 2000 epoch BC-only warmup period. We use an ELU activation function and follow every activation layer with a LayerNorm \citep{ba2016layer}.

For the PADL baseline, we train the model for 14600 epochs, corresponding to approximately 700K optimization steps, 15B frames, and 89 hours of training. For the PADL+BC baseline, we train the model for 14000 epochs, corresponding to approximately 670K optimization steps, 15B frames, and 98 hours of training. We do not use an initial BC-only warmup period (we do not observe it to have a significant impact). For both baselines, we use ReLU activations and no LayerNorm.

\end{document}